\documentclass[pdflatex,sn-mathphys-num]{sn-jnl}% Math and Physical Sciences Numbered Reference Style
\usepackage{graphicx}%

\usepackage{amsmath,amssymb,amsfonts}%
\usepackage{amsthm}%
\usepackage{mathrsfs}%
\usepackage[title]{appendix}%
\usepackage{xcolor}%
\usepackage{textcomp}%
\usepackage{manyfoot}%
\usepackage{booktabs}%
\usepackage{algorithm}%
\usepackage{algorithmicx}%
\usepackage{algpseudocode}%
\usepackage{listings}%
\usepackage{makecell}
\usepackage{pifont}

\usepackage{tabularx}
\usepackage{array}
\usepackage{multirow}%

\usepackage[most]{tcolorbox}
\newcolumntype{Y}{>{\centering\arraybackslash}X}

\tcbuselibrary{listings,breakable}

\newtcblisting{codebox}{
  listing only,
  breakable,
  colback=gray!5,
  colframe=gray!45,
  arc=1.5mm,
  boxrule=0.35pt,
  left=1mm,
  right=1mm,
  top=1mm,
  bottom=1mm,
  listing options={
    basicstyle=\ttfamily\scriptsize,
    breaklines=true,
    columns=fullflexible,
    keepspaces=true
  }
}

\definecolor{conceptualbg}{RGB}{232,242,252}
\definecolor{implementationbg}{RGB}{235,246,232}
\definecolor{debuggingbg}{RGB}{252,235,235}
\definecolor{reflectivebg}{RGB}{242,235,250}
\theoremstyle{thmstyleone}%
\theoremstyle{thmstyletwo}%

\theoremstyle{thmstylethree}%

\begin{document}

\title{AI-Assisted Help-Seeking Trajectories in Programming Education from an SRL-Informed Perspective}


%\author{Anonymous}

\author*[1]{\fnm{Boxuan} \sur{Ma}}\email{boxuan@artsci.kyushu-u.ac.jp}

\author[2]{\fnm{Huiyong} \sur{Li}}

\author[3]{\fnm{Gen} \sur{Li}}

\author[4]{\fnm{Li} \sur{Chen}}

\author[3]{\fnm{Atsushi} \sur{Shimada}}

\author[1]{\fnm{Shin'ichi} \sur{Konomi}}

\affil[1]{\orgdiv{Faculty of Arts and Science}, \orgname{Kyushu University}, \orgaddress{\city{Fukuoka}, \country{Japan}}}

\affil[2]{\orgdiv{Research Institute for Information Technology}, \orgname{Kyushu University}, \orgaddress{\city{Fukuoka}, \country{Japan}}}

\affil[3]{\orgdiv{Faculty of Information Science and Electrical Engineering}, \orgname{Kyushu University}, \orgaddress{\city{Fukuoka}, \country{Japan}}}

\affil[4]{\orgdiv{Division of Math, Sciences, and Information Technology in Education}, \orgname{Osaka Kyoiku University}, \orgaddress{\city{Osaka}, \country{Japan}}}


\abstract{
Generative AI tools provide novice programmers with instant, personalized support, but also raise concerns about whether AI use supports or bypasses students' regulation of problem-solving. Existing work has largely focused on correctness, usability, or overall usage frequency, with less attention to how student--AI help-seeking unfolds. This study addresses this gap by analyzing AI-assisted help-seeking trajectories in university-level programming. 
Using an SRL-informed analytical framework that links prompt-level help-seeking codes to conceptual, implementation, debugging, and reflective forms of support, we analyzed 1,290 task-specific student prompts linked to 17,190 code submissions from 71 students in introductory Python programming courses. Specifically, we examined how help-seeking interactions were structured across turns and attempts, and how trajectory patterns related to task scores and the number of code submissions. Results indicate that many students primarily used AI for reactive troubleshooting rather than for planned, self-regulated problem-solving. Although trajectory patterns were not associated with significant differences in task scores, they differed substantially in the number of code submissions required. These findings suggest that the educational significance of AI support lies not only in whether students use AI, but in how their help-seeking trajectories develop during programming problem-solving. 
}




\maketitle

\section{Introduction}\label{sec1}

Students learning to program routinely encounter deep uncertainty. They may stare at a blank editor, unsure how to begin, struggle to decipher cryptic error messages when code fails, or doubt whether their solution is efficient or aligned with best practices. These moments often arise precisely when help is hardest to access: office hours may not align with students’ schedules, online forums can feel intimidating, and in crowded classrooms, attention often goes to the most vocal students, leaving feedback uneven and delayed \cite{smith2017office,smith2017my}. In higher-education programming courses, large enrollments, wide variation in prior experience, and limited instructional resources further constrain the timely, individualized support that novices need \cite{smith2017office,smith2017my}. Such delays or generic feedback are associated with shallow understanding, weaker retention, limited transfer, and underdeveloped higher-order learning skills \cite{national2018assessing}. Compounding these challenges, many beginners also lack a firm grasp of programming fundamentals, including syntax, control flow, and algorithms \cite{qian2017students}. Addressing this gap requires scalable, equitable, and learner-centered approaches that provide guidance when students encounter difficulty.

Large language models (LLMs) offer a promising avenue for addressing this need. They can instantly generate explanations, examples, and guidance, providing support precisely when students need help \cite{tian2304chatgpt,yilmaz2023augmented}. Recent work suggests that AI tools can augment students’ workflows and make assistance broadly available both in and outside class \cite{kazemitabaar2024codeaid,sun2024would}. These capabilities make AI a compelling tool for narrowing the long-standing gap between when students encounter difficulty and when they receive help. At the same time, this strength also introduces new pedagogical risks. Because AI can often produce working solutions with minimal prompting \cite{denny2023conversing,finnie-ansley2022robots}, students may offload problem-solving steps to AI assistants rather than engage in their own reasoning and problem-solving \cite{anagnostopoulos2023chatgpt,rajala2023call}. Scholars have warned that such use may encourage “metacognitive laziness,” where learners rely on AI to do the thinking for them \cite{darvishi2024impact,fan2025beware}. These concerns extend beyond short-term task completion to issues of academic integrity and the long-term development of independent problem-solving skills \cite{skjuve2023user,tlili2023if}.

Recent research has begun to examine how students use AI tools in programming education, including their prompting behaviors, perceptions of AI support, and the relationship between AI assistance and programming performance \cite{yilmaz2023augmented,ma2024enhancing,zhang2024students}. Much of this work has focused on overall usage patterns and usability \cite{kazemitabaar2024codeaid,sun2024would}. However, describing how often students use AI or what types of prompts they submit provides only a partial view of AI-assisted programming learning. Programming problem-solving is closely related to self-regulated learning (SRL), as students must understand task requirements, plan solutions, monitor errors, revise code, and evaluate outcomes throughout the problem-solving process \cite{loksa2022metacognition}. From this perspective, student--AI help-seeking should not be treated merely as a set of requests for assistance. Similar help-seeking actions may reflect different forms of cognitive and regulatory engagement: some students may seek AI support after diagnosing a problem and monitoring their progress, whereas others may repeatedly seek fixes without articulating a diagnosis. Therefore, student--AI interaction needs to be examined not only in terms of prompt types or usage frequency, but also as trajectories of help-seeking behavior that may reveal how students regulate programming problem-solving. In particular, we know relatively little about what kinds of help-seeking trajectories students naturally develop when interacting with AI during programming tasks and how these trajectories are associated with student performance \cite{darvishi2024impact,fan2025beware,prather2023robots}.

To address this gap, we examine AI-assisted help-seeking trajectories in university-level programming from an SRL-informed perspective \cite{zimmerman2000attaining,loksa2022metacognition}. We position this work within the broader literature on student--AI interaction, but focus specifically on task-related interactions in which students sought AI support during programming exercises. These interactions are conceptualized as AI-mediated help-seeking episodes that may differ in their regulatory orientation, ranging from conceptual clarification to implementation, debugging, verification, and reflection-oriented support. This framing allows us to examine how students' task-related AI interactions unfold during programming problem solving, how these trajectories relate to programming performance, and what regulatory opportunities or risks they may reflect \cite{fan2025beware,phung2025plan}.

Guided by this motivation, our study addresses the following research questions:
\begin{itemize}

 \item \textbf{RQ1 - Help-Seeking Trajectories:} How do students’ help-seeking trajectories unfold when they interact with AI assistants during programming problem solving?
 
 \item \textbf{RQ2 - Interaction Patterns and Programming Performance:} What attempt-level interaction patterns emerge in student--AI help-seeking sessions, and how are these patterns associated with task scores and the number of code submissions?

 \item \textbf{RQ3 - SRL-Informed Interpretation:} How can these help-seeking trajectories and interaction patterns be interpreted from an SRL-informed perspective?

\end{itemize}

By examining how help-seeking unfolds across turns, attempts, and tasks, this paper contributes an SRL-informed analytical framework for interpreting student--AI help-seeking, identifies recurring interaction trajectories, and links these trajectories to programming performance. In doing so, it provides empirical insight into when AI-supported interaction may scaffold programming problem-solving and when it may instead be associated with reactive troubleshooting and higher process cost.

\section{Related Work}

\subsection{Generative AI in Programming Education}

Generative AI has rapidly become part of programming education by lowering barriers to help-seeking and supporting a wide range of tasks, including code generation, debugging, code explanation, and personalized feedback \cite{rajala2023call,denny2023conversing}. Early work in this area primarily examined the capabilities of generative AI tools on programming tasks, showing that they could solve many common CS1 and CS2 problems and, in some cases, achieve near-human or above-median student performance \cite{finnie-ansley2022robots,finnie2023my,phung2023generative,sarsa2022automatic,savelka2023thrilled}. Other studies further showed that these tools can provide clear explanations and useful feedback that support learners’ understanding, although concerns remain about inaccuracy, overconfidence, and the pedagogical quality of generated responses \cite{macneil2023experiences,leinonen2023comparing,pankiewicz2023large,zhang2024students}.

As these tools entered authentic learning contexts, research increasingly shifted from what AI tools \emph{can} do to how students and educators \emph{use} and \emph{perceive} them in programming education \cite{biswas2023role,humble2023cheaters,yilmaz2023augmented,shoufan2023exploring,ma2024enhancing,ma2024exploring}. Previous studies have shown that students often value AI for their immediacy, accessibility, and clarity, especially when compared with searching online resources or waiting for human support \cite{biswas2023role,yilmaz2023augmented,ma2024enhancing}. At the same time, they remain aware that AI-generated answers may be incorrect or misleading, and that over-reliance may weaken deeper understanding, persistence, and critical thinking \cite{shoufan2023exploring,skjuve2023user,tlili2023if,kasneci2023chatgpt}. Educators likewise recognize the scalability and convenience of AI-based support, but often express stronger concerns about over-reliance, misuse, academic integrity, and the need for appropriate pedagogical and institutional responses \cite{gao2022who,denny2023computing,chan2023ai,amani2023generative,prather2023robots,lau2023ban,becker2023programming}.

\subsection{Student--AI Interaction in Programming Education}

A growing set of studies has shifted to examine students' prompts and broader usage patterns in authentic learning settings \cite{ma2024enhancing,ma2024exploring,amoozadeh2024student,scholl2024chatprotocols,scholl2024novice}. These studies show that student--AI interaction is not uniform: learners use AI for conceptual explanation, code generation, debugging, refinement, and confirmation, and they differ in how much context they provide, how they follow up on AI responses, and how critically they evaluate the returned output \cite{amoozadeh2024student,scholl2024chatprotocols,scholl2024novice}.

Several studies have examined student--AI interaction in programming as a form of help-seeking \cite{sheese2024patterns,amoozadeh2024student,viberg2025chatting}. Prior work has shown that students' requests are often oriented toward immediate assignment needs, debugging, and code completion \cite{scholl2024chatprotocols,scholl2024novice,yang2024debugging}. Other studies further suggest that scaffolded AI systems, prompting support, and structured integration can shape the quality of student--AI interaction \cite{liffiton2023codehelp,kazemitabaar2024codeaid,sun2024would,denny2023conversing}. At the same time, critical work warns that easy access to AI-generated answers can encourage over-reliance, reduce learner agency, or amplify differences between productive and shallow forms of AI use \cite{darvishi2024impact,prather2024widening,fan2025beware}. Together, these studies suggest that the central issue is not simply whether students use AI, but how their interactions unfold and what kinds of help-seeking processes those interactions reflect.

\subsection{Help-Seeking and Self-Regulated Learning in Programming Education}

Self-regulated learning and help-seeking have long been recognized as important in programming education, where learners must not only acquire technical knowledge but also regulate their problem-solving approach, monitor their progress, and evaluate their solutions \cite{loksa2022metacognition}. Novice programmers often struggle to structure solution strategies, identify and resolve errors, and judge the quality or efficiency of their work \cite{ebrahimi2006taxonomy,hao2025towards,park2025exploring,saliba2024learning}. Prior research has shown that support for reflection and regulation can improve learning processes and outcomes. For example, Choi et al. \cite{choi2023benefit} found that prompting reflection after programming tasks improved performance in both immediate and delayed tests. Similarly, Yilmaz and Yilmaz \cite{karaoglan2022learning} reported higher engagement among students who received personalized feedback, and Cheng et al. \cite{cheng2024exploring} showed that high-performing students relied more on elaboration and critical thinking, whereas lower-performing peers used more basic strategies. Taken together, this body of work suggests that programming learning should be understood not only in terms of task completion but also in terms of how learners regulate their help-seeking and problem-solving processes.

The emergence of generative AI introduces a new context for these processes, as AI can now serve as an always-available source of explanations, suggestions, and solutions. Recent work has therefore raised concerns that easy access to AI-generated answers may reduce learners’ engagement in reflection and regulation, a phenomenon described as ``metacognitive laziness'' \cite{fan2025beware,ma2025scaffolding}. Related studies suggest that learners may bypass evaluative or monitoring processes when interacting with AI-powered tools \cite{chen2025unpacking}, even though scaffolded forms of support, such as step-by-step pseudo-code or constrained feedback, may encourage more reflective engagement \cite{kazemitabaar2024codeaid,kazemitabaar2023studying}. Other recent evidence further suggests that students do not treat AI as a neutral replacement for other help resources. Instead, they weigh factors such as trust, convenience, and task stakes when deciding whether and how to seek help from AI \cite{penney2025preferences,li2025coderunner}.

However, while prior work has established the importance of self-regulation in programming learning and raised concerns about AI-supported learning, much less is known about how these processes are reflected in students’ help-seeking interactions with AI. In particular, we still know relatively little about how help-seeking unfolds across turns and attempts when students interact with AI during programming problem-solving. For this reason, we use self-regulated learning as an interpretive lens for understanding AI-assisted help-seeking trajectories.

\section{Method}

We conducted a study in an introductory undergraduate Python programming course at a national university in Japan. The study comprised two offerings of the same course, one in summer 2024 and one in summer 2025, with a total of 166 students (112 in 2024 and 54 in 2025). Throughout each course, all students were given access to an AI tool as a supplementary learning resource. Access was provided via a custom web-based interface connected to GPT-4o, enabling students to interact with the model in real time. All student--AI conversations through this interface were systematically logged for analysis. The study was approved by the university’s ethics review board prior to its commencement.

\subsection{Course Context and Participants}

\subsubsection{Course Structure}

This first-year undergraduate course, designed for beginners, covered the foundations of the Python programming language in 14 lessons delivered over one semester. Each 90-minute lesson was divided into two segments: 45 minutes of direct instruction supported by lecture presentations and slides, followed by 45 minutes devoted to programming tasks. This structure was intended to enable students to apply the theoretical concepts introduced in class.

Both course offerings were taught by the same instructor. Over the semester, students completed 57 programming tasks. These tasks were organized into weekly assignments, each of which typically contained 3--8 tasks, averaging 4.75 tasks per lesson. The tasks covered introductory Python topics, including variables, strings, functions, conditional statements, and loops. During the programming task segment, students completed these tasks using CodeRunner, a Moodle plugin for automated programming assessment that allows students to write and submit code within the learning management system. In addition to the AI tool, students had access to lecture slides and support from the course instructors.

\subsubsection{Participants}

The study involved 166 undergraduate students across two offerings of the same course, most of whom were first-year students, with 60.2\% male and 39.8\% female participants. Participation was voluntary, informed consent was obtained from all participants, and no aspect of participation affected students’ course grades.

\subsection{Data Sources and Sample Construction}

\subsubsection{Data Sources}

We drew on two complementary trace data sources: CodeRunner submission records and student--AI dialogue logs. Together, these sources enabled us to examine not only how students interacted with AI during programming problem-solving, but also how those interactions relate to concrete task attempts and their performance.

\paragraph{Automated Assessment Submissions.} 

Programming tasks were delivered and assessed using CodeRunner, a Moodle plugin for automated programming assessment. For each task, students could read the problem statement, enter their code, and submit their solutions for automatic checking against pre-defined test cases. Each submission was timestamped, logged, and scored automatically by the system. Students could submit multiple times as they worked toward a correct solution, and the submission logs therefore capture both successful and unsuccessful attempts during problem solving. Task scores were assigned based on the number of test cases passed: submissions that passed all test cases received full credit, submissions that passed a subset of test cases received partial credit, and submissions that failed all test cases received no credit.

These submission records served as behavioral traces of students' task progress. The original records contained 37,558 code submissions across the two course offerings. After applying the analytic-sample filter, the 71 students included in the analysis were linked to 17,190 code submissions. This corresponded to an average of 4.1 submissions per student-task pair. Because students could submit multiple times, the overall solve rate was high (0.98), with only a small number of student-task pairs remaining unsolved.

\paragraph{Dialogue Logs.}

Another data source for this study was student--AI dialogue logs, which capture students’ interactions with AI \cite{kazemitabaar2023studying,kazemitabaar2023how_novices_use_llm_code} and provide direct evidence of how learners seek and use support during programming problem-solving \cite{brandt2009two}. In this study, we used dialogue logs to examine students’ help-seeking interactions with AI and how these interactions related to concrete programming attempts.

Across the two course offerings, 166 students were enrolled. For the present analysis, we included students who generated at least five prompts during the course, ensuring their dialogue data provided sufficient information for task-level analysis. This filtering procedure yielded an analytic sample of 71 students, comprising 30 from the 2024 cohort and 41 from the 2025 cohort. These students were linked to 17,190 code submissions. Students who did not use AI through the course interface or contributed fewer than five prompts were excluded from the analysis. The resulting dataset contained 3,649 raw dialogue prompts from 744 conversations. Each conversation corresponded to one student’s interaction with AI during a weekly assignment.

\subsubsection{Task-Level Segmentation and Sample Construction}

To focus on interactions that occurred during problem solving, we excluded prompts unrelated to the course and those submitted outside the exercise period. After filtering, the analytic sample consisted of 1,290 task-specific prompts.

To align interaction data more closely with specific problem-solving episodes, we segmented each conversation by task. This procedure yielded 635 task-specific dialogue sessions, comprising the 1,290 prompts analyzed. These sessions were then linked to the corresponding submission logs, enabling us to examine how help-seeking unfolded within task attempts and how interaction patterns related to student performance. Table~\ref{tab:analytic_sample} summarizes the analytic sample and units of analysis.

\begin{table}[t]
\centering
\caption{Summary of the analytic sample and units of analysis.}
\label{tab:analytic_sample}
\small
\setlength{\tabcolsep}{6pt}
\renewcommand{\arraystretch}{1.15}
\begin{tabularx}{\linewidth}{
    >{\raggedright\arraybackslash}p{0.30\linewidth}
    >{\raggedright\arraybackslash}p{0.50\linewidth}
    >{\centering\arraybackslash}X
}
\toprule
\textbf{Section} & \textbf{Measure} & \textbf{Value} \\
\midrule
\multirow{3}{*}{Students}
& Students in analytic sample & 71 \\
& 2024 class& 30 \\
& 2025 class& 41 \\
\midrule
\multirow{3}{*}{\makecell[l]{Dialogue Data}}
%& Conversations & 744 \\
%& Raw dialogue prompts & 3,649 \\
& Task-specific sessions & 635 \\
& Task-specific prompts & 1,290 \\
\midrule
\multirow{3}{*}{\makecell[l]{Tasks and Submissions}}
& Programming tasks & 57 \\
& Code submissions & 17,190 \\
& Average submissions per student-task pair & 4.1 \\
& Solve rate & 0.98 \\
\bottomrule
\end{tabularx}
\end{table}

\begin{table}[t]
\centering
\caption{Help-seeking classification scheme with prompt codes.}
\label{tab:codebook}
\small
\setlength{\tabcolsep}{6pt}
\renewcommand{\arraystretch}{1.15}
\begin{tabularx}{\textwidth}{
    >{\centering\arraybackslash}p{0.15\textwidth}
    >{\raggedright\arraybackslash}p{0.33\textwidth}
    >{\raggedright\arraybackslash}X
}
\toprule
\textbf{Category} & \textbf{Code} & \textbf{Description} \\
\midrule
\multirow{3}{0.15\textwidth}{\centering Conceptual\\Help}
& Concept Question (CQ) & Asking about programming concepts or syntax \\
& Problem Understanding (PU) & Asking about the meaning of code, expressions, or task requirements \\
& Example Request (ER) & Requesting concrete examples \\
\midrule
\multirow{2}{0.15\textwidth}{\centering Implementation\\Help}
& Implementation Question (IQ) & Asking how to implement specific functionality \\
& Code Generation (CG) & Providing a problem description and requesting code \\
\midrule
\multirow{3}{0.15\textwidth}{\centering Debugging\\Help}
& Error Interpretation (EI) & Sharing an error message and asking for interpretation \\
& Code Correction (CC) & Sharing broken code and asking for correction \\
& Code Verification (CV) & Sharing code and asking for verification \\
\midrule
\multirow{2}{0.15\textwidth}{\centering Reflective\\Help}
& Code Explanation (CE) & Asking for an explanation of working code \\
& Code Optimization (CO) & Asking for better or alternative approaches \\
\bottomrule
\end{tabularx}
\end{table}

\subsection{Help-Seeking Coding}

Because our analysis focused only on task-specific student--AI interactions during programming exercises, we conceptualized these prompts as AI-mediated help-seeking actions. This framing is consistent with previous work that analyzed student prompts in programming education through task-related and SRL-informed categories \cite{viberg2025chatting}. Here, help-seeking refers broadly to students' attempts to obtain conceptual, procedural, debugging, or reflective support from AI while working on programming tasks. This definition allows us to examine task-specific prompts as different forms of support-seeking during programming problem-solving.

\subsubsection{Codebook}

Our coding unit was the \emph{task-specific prompt}, defined as a student prompt for a concrete programming task. Drawing on prior work on AI-supported programming and self-regulated help-seeking \cite{kazemitabaar2024codeaid,ma2024exploring,karabenick2013help,loksa2022metacognition,prasad2024self,silva2024learning}, we organized student prompts according to the functional role of the requested help in programming problem-solving. The resulting codebook contains 10 fine-grained codes organized under four main help-seeking categories: \emph{Conceptual}, \emph{Implementation}, \emph{Debugging}, and \emph{Reflective} (Table~\ref{tab:codebook}). These four categories allow us to characterize the functional role of help-seeking prompts in programming problem-solving while retaining sufficient granularity for sequential analysis. 

\emph{Conceptual Help} captures requests to build understanding before or alongside coding. These prompts focus on clarifying programming concepts, interpreting code or expressions, and requesting concrete examples. This category includes \emph{Concept Question} (CQ), \emph{Problem Understanding} (PU), and \emph{Example Request} (ER). 

\emph{Implementation Help} captures requests to construct a solution or generate code to move the task forward. These prompts focus on implementing specific functionality or directly asking the AI to produce code. This category includes \emph{Implementation Question} (IQ) and \emph{Code Generation} (CG). 

\begin{figure}[t]
  \centering
  \includegraphics[width=\linewidth]{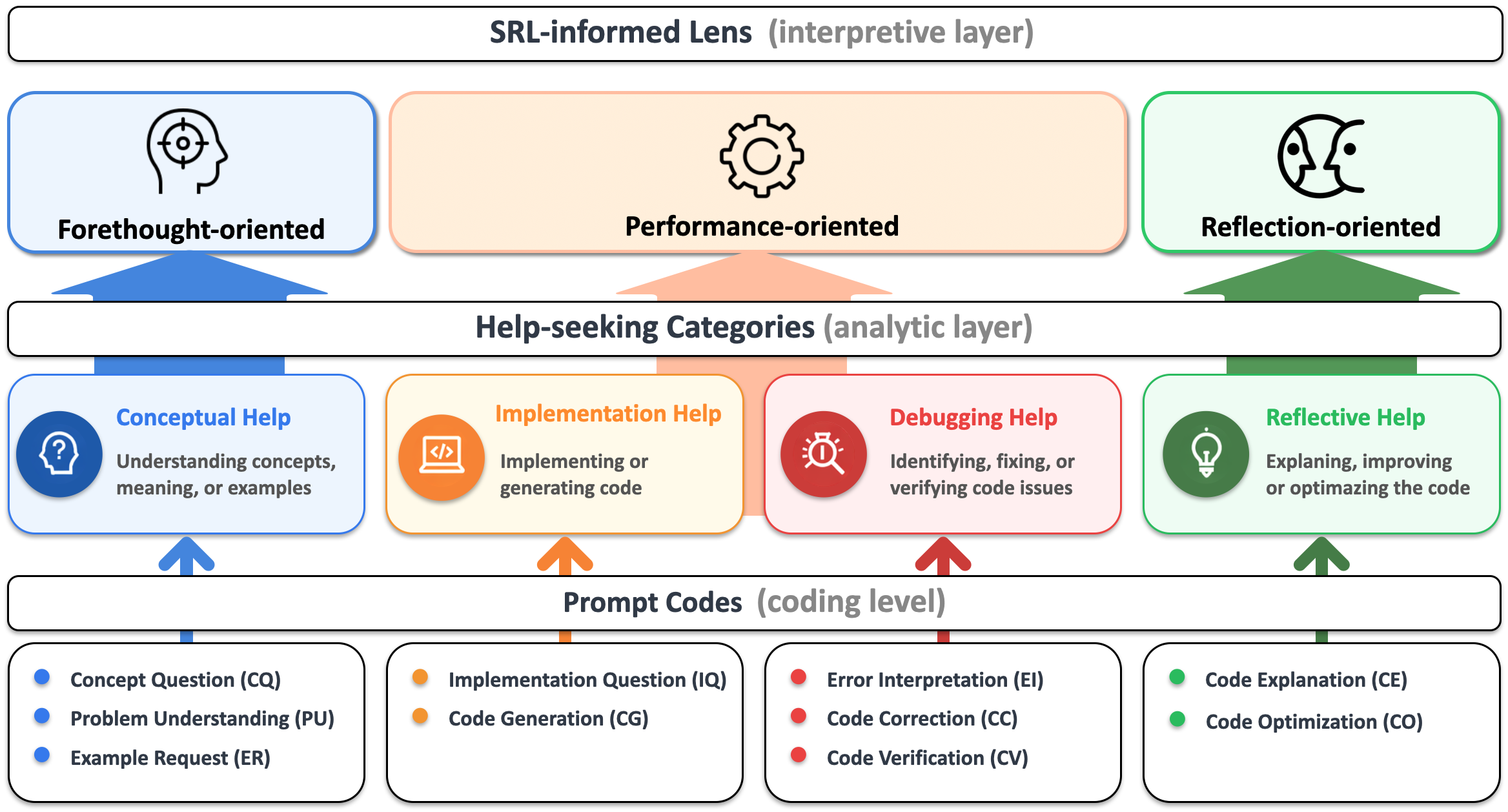}
\caption{Three-layer analytical framework used in this study. Prompt codes are first assigned to task-specific prompts, then aggregated into four help-seeking categories, and finally interpreted through an SRL-informed lens.}
  \label{fig:framework}
\end{figure}

\emph{Debugging Help} captures requests that arise when students encounter problems during code execution or verification. These prompts focus on interpreting error messages, fixing broken code, or checking whether code behaves as intended. This category includes \emph{Error Interpretation} (EI), \emph{Code Correction} (CC), and \emph{Code Verification} (CV). 

\emph{Reflective Help} captures requests that go beyond immediate task completion and instead focus on explaining, improving, or refining a working solution. These prompts concern understanding why code works, considering alternative approaches, and improving code quality. This category includes \emph{Code Explanation} (CE) and \emph{Code Optimization} (CO).

To clarify how the prompt-level codes are organized analytically and interpreted theoretically, Figure~\ref{fig:framework} presents the three-layer framework used in this study. At the bottom layer, task-specific prompts are assigned one of 10 fine-grained help-seeking codes. These codes are then grouped into four broader help-seeking categories: Conceptual, Implementation, Debugging, and Reflective. At the interpretive layer, these categories are understood through an SRL-informed lens. Conceptual help aligns with forethought-oriented support, Implementation and Debugging are both situated within performance-oriented processes, and Reflective help aligns with reflection-oriented support. Importantly, the SRL layer is used as a theoretical interpretation of observed help-seeking patterns rather than as a direct coding dimension.

\subsubsection{Coding Procedure and Validation}

Following an inductive coding approach \cite{bingham2021deductive}, two researchers jointly reviewed an initial randomly sampled subset comprising approximately 30\% of the prompt pool to generate preliminary codes. They then independently coded an additional subset of approximately 5\% of the prompts using this preliminary codebook. These results were discussed with the course instructors, and disagreements were resolved through consensus, yielding a refined version of the codebook. To establish reliability, the two researchers independently coded a further subset of approximately 10\% of the prompts, and inter-rater agreement was assessed using Cohen’s kappa and percentage agreement \cite{miles1994qualitative,neuendorf2017content}. After satisfactory reliability was achieved, the refined codebook was applied to the remaining dataset. The final inter-rater agreement was 92.7\%, with a Cohen’s $\kappa$ of .88. The two researchers reviewed all remaining disagreements, discussed the rationale for each code, and resolved them through consensus. The consensus-coded dataset was then used for the final analyses.

\section{Results}

\subsection{Distribution of Help-Seeking Prompts}

\begin{figure}[t]
  \centering
  \includegraphics[width=0.7\linewidth]{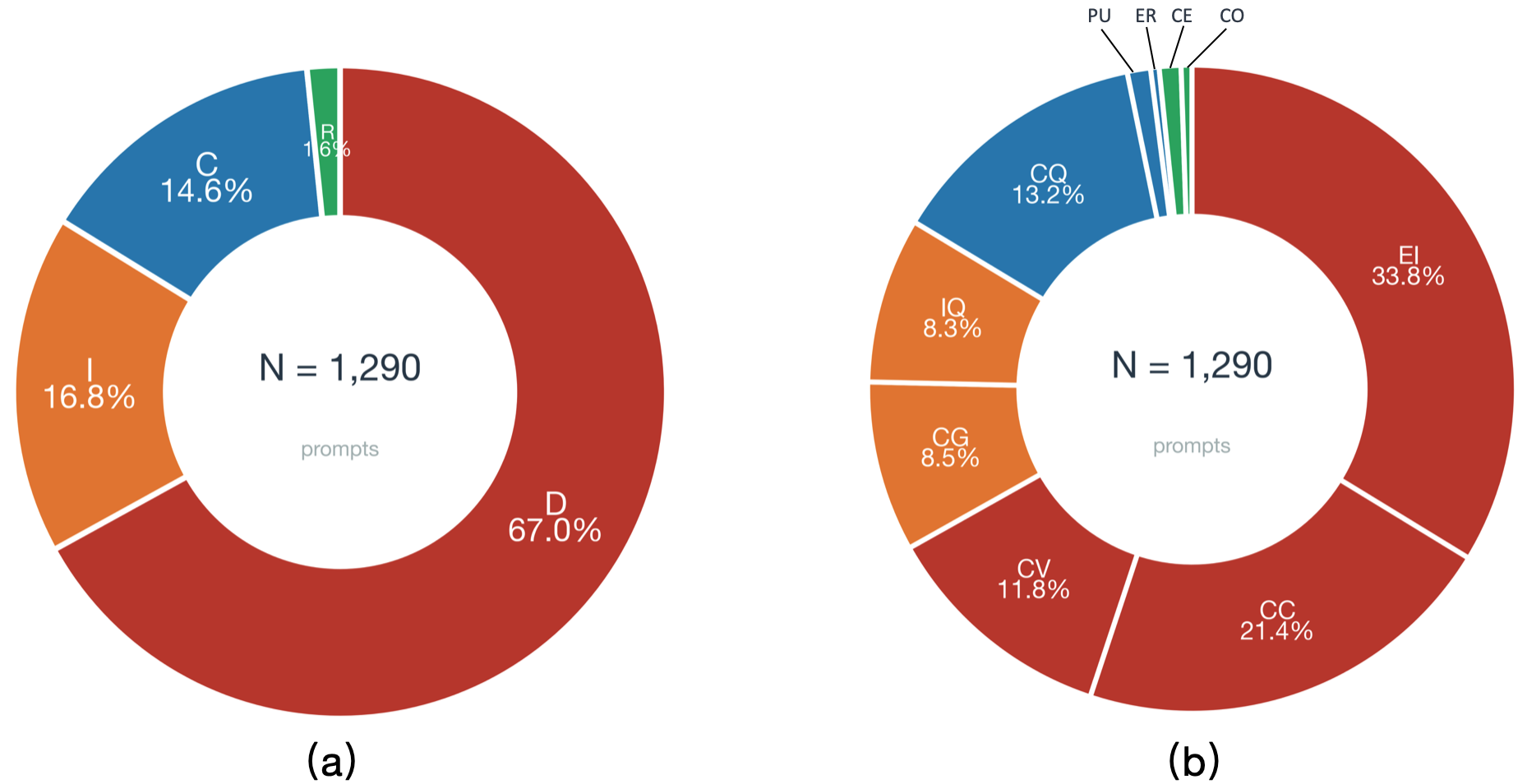}
\caption{Distribution of help-seeking prompts at the category and code levels. Panel (a) shows the four main help-seeking categories, and panel (b) shows the corresponding prompt codes.}
  \label{fig:distribution}
\end{figure}

We first examined the overall distribution of help-seeking prompts across the four main categories and their fine-grained codes. As shown in Figure~\ref{fig:distribution}(a), Debugging (D) was by far the most common category, accounting for 67.0\% of all prompts ($n = 864$). In comparison, Implementation (I) accounted for 16.8\% ($n = 217$), Conceptual (C) for 14.6\% ($n = 188$), and Reflective (R) for only 1.6\% ($n = 21$). 

Figure~\ref{fig:distribution}(b) further breaks down these categories into fine-grained prompt codes. The most frequent code was Error Interpretation (EI), accounting for 33.8\% of all prompts ($n = 436$). This was followed by Code Correction (CC) at 21.4\% ($n = 276$), Concept Question (CQ) at 13.2\% ($n = 170$), and Code Verification (CV) at 11.8\% ($n = 152$). Within the Implementation category, Code Generation (CG) and Implementation Question (IQ) were relatively balanced, accounting for 8.5\% ($n = 110$) and 8.3\% ($n = 107$), respectively. By contrast, the remaining codes were rare: Problem Understanding (PU) accounted for 1.2\% ($n = 15$), Code Explanation (CE) for 1.1\% ($n = 14$), Code Optimization (CO) for 0.5\% ($n = 7$), and Example Request (ER) for only 0.2\% ($n = 3$).

Taken together, these results suggest that students primarily used AI as a reactive programming aid. Most prompts occurred when students encountered concrete coding problems, particularly errors or incorrect outputs. In contrast, prompts aimed at conceptual clarification, example-based exploration, explanation, or optimization appeared much less frequently. This distribution provides the descriptive basis for the subsequent sequential analyses, which examine whether these help-seeking modes appeared as isolated requests or formed persistent trajectories across turns.

\begin{figure}[t]
  \centering
  \includegraphics[width=\linewidth]{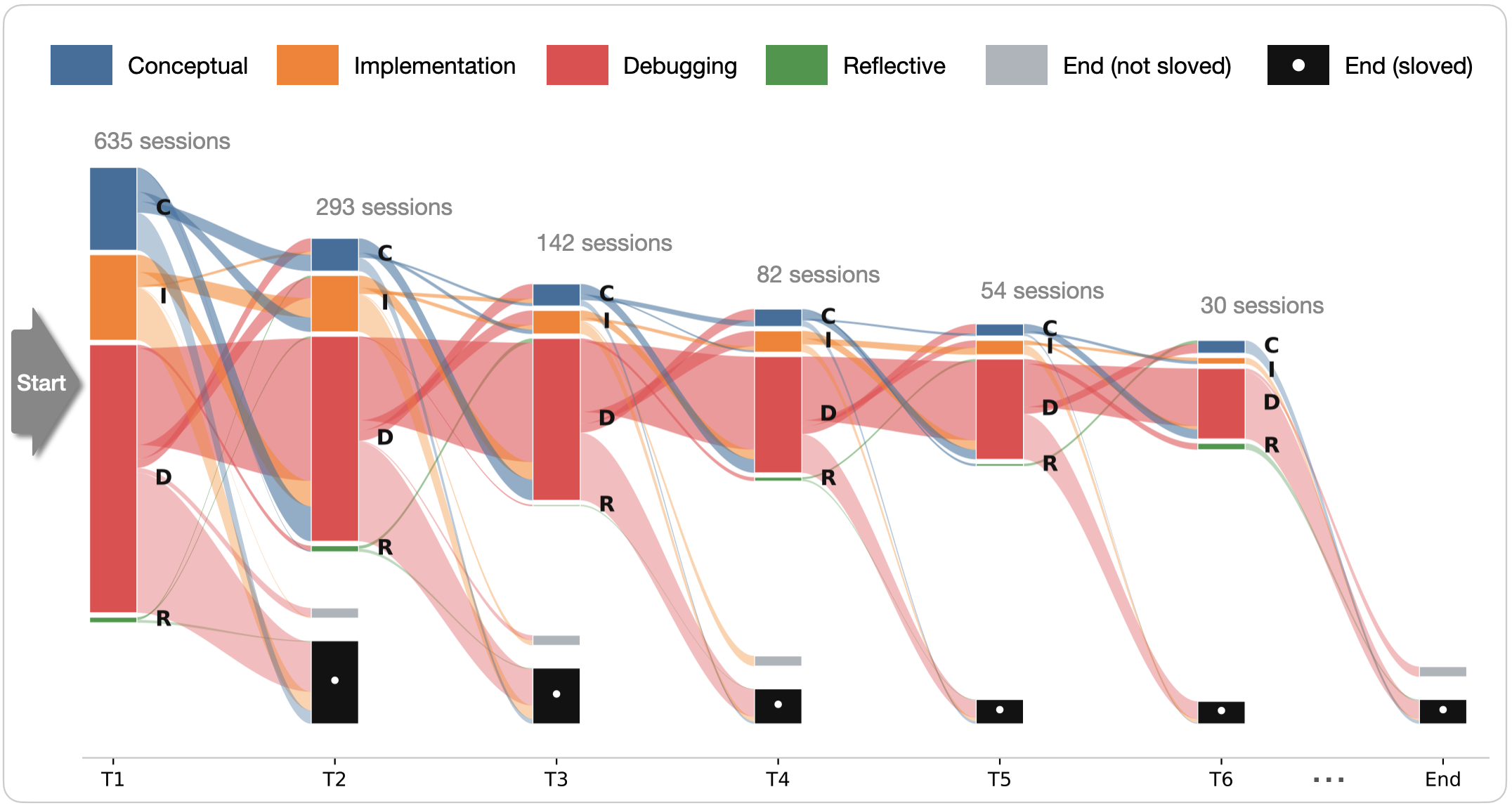}
\caption{Category-level help-seeking trajectories across task-specific sessions. Each column represents one turn in the interaction sequence (T1-T6), and the width of each flow indicates the number of sessions following that transition. Colored bars show the four main help-seeking categories: Conceptual (C), Implementation (I), Debugging (D), and Reflective (R). The black and grey bars indicate whether sessions ended with a solved or unsolved outcome.}
  \label{fig:sankey}
\end{figure}

\subsection{Sequential Patterns of Help-Seeking Prompts}

Figure~\ref{fig:sankey} visualizes the category-level prompt sequences as a Sankey diagram. Each column represents one turn in the interaction sequence (T1-T6), and the width of each flow indicates the number of sessions following that transition. The figure shows three main patterns. First, Debugging dominated from the first turn and remained the largest stream throughout later turns, indicating that many students entered AI interaction through error interpretation, code correction, or verification. Second, many sessions were short, with the number of active sequences dropping substantially after the first turn. Third, Reflective prompts appeared only as thin streams, suggesting that students rarely used AI for explanation, improvement, or optimization during task-specific problem solving.

\begin{figure}[t]
  \centering
  \includegraphics[width=\linewidth]{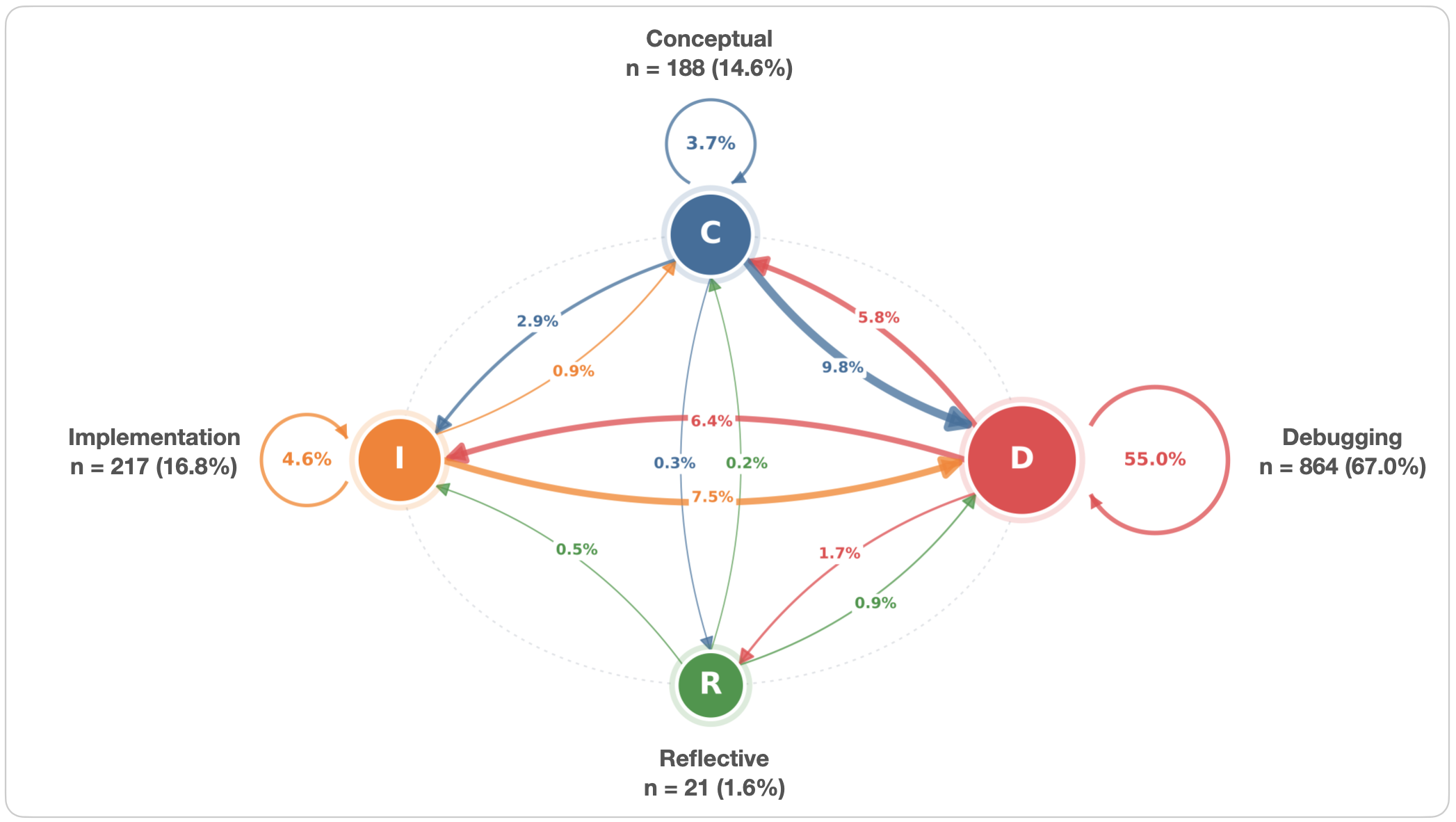}
\caption{Category-level transition network across task-specific sessions. Nodes represent the four main help-seeking categories and are sized by the overall frequency of prompts. Directed edges indicate the percentage of within-attempt transitions flowing between categories, with arrow thickness scaled by transition frequency. Self-loops represent within-category persistence.}
  \label{fig:tna}
\end{figure}

Figure~\ref{fig:tna} complements this turn-by-turn view by showing the category-level transition network. Debugging occupied the most central position in the network, both in overall prompt volume and in persistence. It was the largest node ($n = 864$, 67.0\%) and had the strongest self-loop (55.0\%), indicating that successive prompts often remained within debugging-oriented help-seeking. Conceptual and Implementation help were also connected to Debugging, suggesting that students sometimes moved among understanding, implementation, and debugging within the same attempt. However, Reflective help remained peripheral, with few incoming or outgoing transitions and little persistence. 

\begin{figure}[t]
  \centering
  \includegraphics[width=0.8\linewidth]{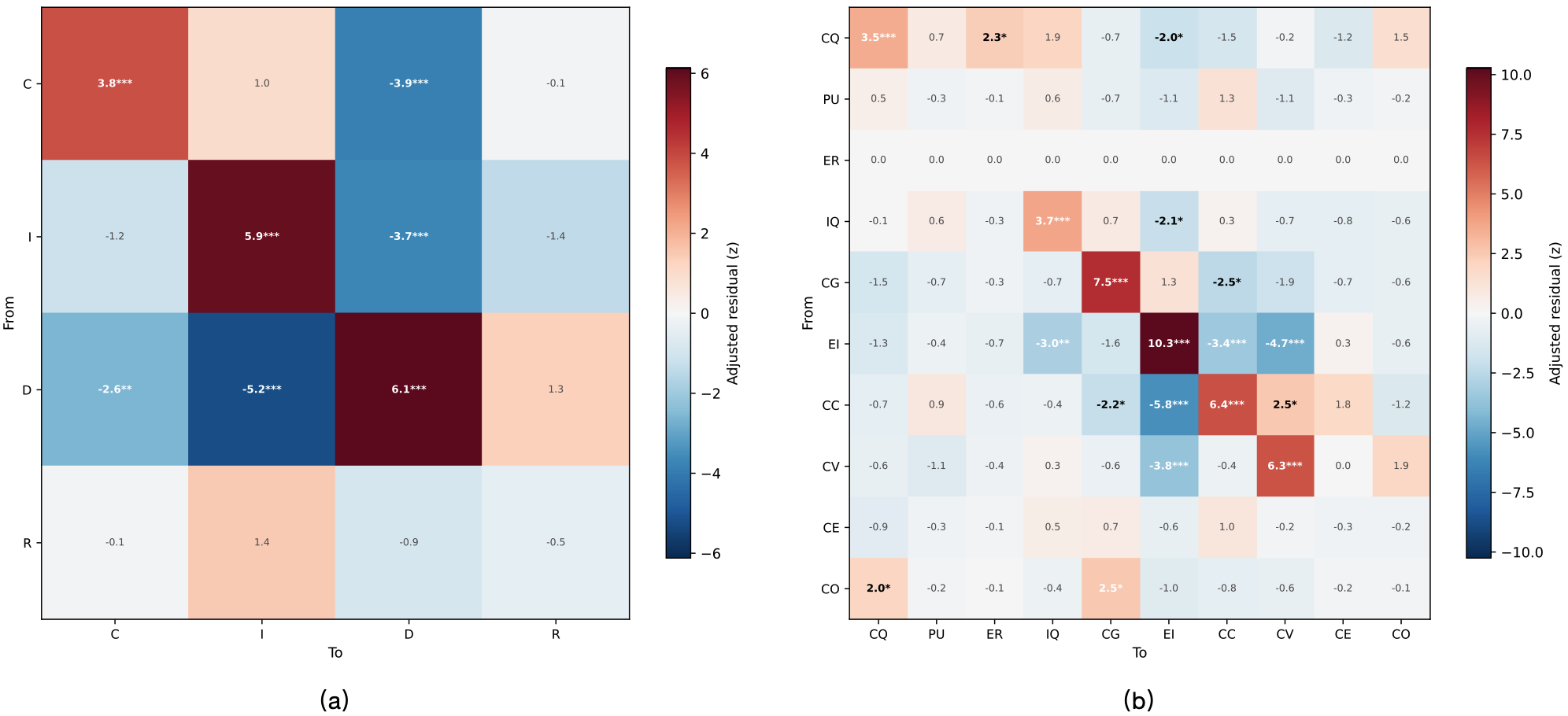}
\caption{Lag sequential analysis of help-seeking transitions. Panel (a) shows adjusted residuals ($z$-scores) for transitions among the four main help-seeking categories: Conceptual (C), Implementation (I), Debugging (D), and Reflective (R). Panel (b) shows adjusted residuals for transitions among the fine-grained prompt codes. Positive values indicate transitions that occurred more often than expected under independence, whereas negative values indicate transitions that occurred less often than expected. Asterisks mark statistically notable cells (* $p<.05$, ** $p<.01$, *** $p<.001$).}
  \label{fig:lsa}
\end{figure}

To test whether these transitions deviated from chance, we conducted lag sequential analysis at both the category and fine-code levels. As shown in Figure~\ref{fig:lsa}(a), category-level self-transitions were significantly elevated for Conceptual ($z = 3.8$), Implementation ($z = 5.9$), and especially Debugging ($z = 6.1$). At the same time, several cross-category transitions were significantly suppressed, including Conceptual~$\rightarrow$~Debugging ($z = -3.9$), Implementation~$\rightarrow$~Debugging ($z = -3.7$), and Debugging~$\rightarrow$~Implementation ($z = -5.2$). 

A similar pattern appeared at the fine-code level. Several code-level self-transitions were strongly elevated, including CQ~$\rightarrow$~CQ ($z = 3.5$), IQ~$\rightarrow$~IQ ($z = 3.7$), CG~$\rightarrow$~CG ($z = 7.5$), EI~$\rightarrow$~EI ($z = 10.3$), CC~$\rightarrow$~CC ($z = 6.4$), and CV~$\rightarrow$~CV ($z = 6.3$). Within debugging-related help-seeking, transitions across subtypes were often suppressed, such as EI~$\rightarrow$~CC ($z = -3.4$), EI~$\rightarrow$~CV ($z = -4.7$), CC~$\rightarrow$~EI ($z = -5.8$), and CV~$\rightarrow$~EI ($z = -3.8$). 

Taken together, these results suggest that students' AI-assisted help-seeking often remained concentrated around immediate troubleshooting, rather than moving flexibly across different forms of support within programming problem-solving.

\subsection{Attempt-Level Taxonomy and Programming Performance}

While the transition analyses clarify dependencies between adjacent prompts, they do not capture the overall organization of a task-specific help-seeking session. For RQ2, we therefore moved from transition-level analysis to attempt-level patterning. This step allowed us to summarize each task-specific session as a single trajectory and examine whether distinct trajectory structures were associated with programming performance, measured by task scores and the number of code submissions.

\subsubsection{Constructing the Attempt-Level Taxonomy}

Each task-specific session was represented as a sequence of the four main help-seeking categories: Conceptual (C), Implementation (I), Debugging (D), and Reflective (R). For example, a conceptual request followed by a debugging request was represented as C--D, whereas three consecutive debugging requests were represented as D--D--D.

Following trace-based SRL and help-seeking research that use theoretically grounded rules or computational models to translate low-level learning actions into interpretable regulatory or help-seeking patterns~\cite{fan2022improving,aleven2006toward}, we developed a theory-informed, rule-based taxonomy to classify sequences of Conceptual, Implementation, Debugging, and Reflective help-seeking into attempt-level trajectory patterns. Before defining the final rules, we used exploratory sequence clustering as a data-driven inspection step to assess whether recurring trajectory structures emerged from the sequence data.

Table~\ref{tab:taxonomy_rules} summarizes the hierarchical decision rules used in the taxonomy construction. The rules were applied in priority order so that dominant structural features were classified before more general sequence characteristics. The rules were applied in the following order:

\begin{table}[t]
\centering
\caption{Hierarchical decision rules for the attempt-level taxonomy.}
\label{tab:taxonomy_rules}
\small
\renewcommand{\arraystretch}{1.15}
\setlength{\tabcolsep}{5pt}
\begin{tabular}{p{0.25\linewidth} p{0.1\linewidth} p{0.57\linewidth}}
\toprule
\textbf{Pattern} & \textbf{Priority} & \textbf{Operational rule} \\
\midrule
One-shot & 1 & The task-specific session contains only one prompt. \\

Debugging-Persistence & 2 & The session contains three or more consecutive Debugging prompts (\texttt{D-D-D}), regardless of opening category. This dominant structural feature overrides later rules. \\

Conceptual-Framing & 3 & The session begins with Conceptual help (\texttt{C}) and does not meet the Debugging-Persistence criterion. \\

Performance-Oriented & 4 & The session consists only of Implementation and Debugging help (\texttt{I} and \texttt{D}), without any Conceptual or Reflective prompts, and does not meet the Debugging-Persistence criterion. \\

Mode-Shifting & 5 & All remaining multi-prompt sessions involving shifts across help-seeking modes, especially sequences that move into Conceptual or Reflective help after an initial Implementation or Debugging state. \\
\bottomrule
\end{tabular}
\end{table}

\begin{enumerate}

\item \textbf{\emph{One-shot}}: Sessions containing only one task-specific prompt were classified as \emph{One-shot}. Although these sessions do not contain within-attempt transitions, they were retained as a separate category because they represented the dominant form of task-specific AI-assisted help-seeking. These sessions could take four forms depending on the prompt category: a single Conceptual, Implementation, Debugging, or Reflective request.

   \item \textbf{\emph{Debugging-Persistence}}: Among multi-prompt sessions, sequences containing three or more consecutive Debugging prompts were classified as \emph{Debugging-Persistence}. This operational threshold was used to capture sustained persistence in the same debugging-oriented mode, consistent with prior help-seeking research that treats repeated unproductive help use or trial-and-error behavior as a meaningful regulatory problem~\cite{aleven2006toward,aleven2016help}. This rule was prioritized because prolonged debugging persistence was treated as the dominant structural feature of the attempt. Thus, sequences such as C--D--D--D or I--D--D--D were classified as \emph{Debugging-Persistence}.
   
\item \textbf{\emph{Conceptual-Framing}}: The remaining sequences that began with Conceptual help were classified as \emph{Conceptual-Framing}. These attempts began with conceptual help, then either shifted to another help-seeking mode or continued with conceptual clarification. This pattern aligns with recent AI-assisted programming work that emphasizes the value of planning-oriented metacognitive support before debugging or implementation~\cite{phung2025plan}.

    \item \textbf{\emph{Performance-Oriented}}: Sequences composed only of Implementation and Debugging help, without Conceptual or Reflective helps and without meeting the \emph{Debugging-Persistence} criterion, were classified as \emph{Performance-Oriented}. These attempts primarily involved task execution, code construction, or short debugging-oriented exchanges. 

\item \textbf{\emph{Mode-Shifting}}: All remaining multi-prompt sequences involving shifts across help-seeking modes, especially transitions into Conceptual or Reflective help after an initial Implementation or Debugging state, were classified as \emph{Mode-Shifting}.

\end{enumerate}

\subsubsection{Distribution of Attempt-Level Trajectory Patterns}

\begin{figure}[t]
  \centering
  \includegraphics[width=1\linewidth]{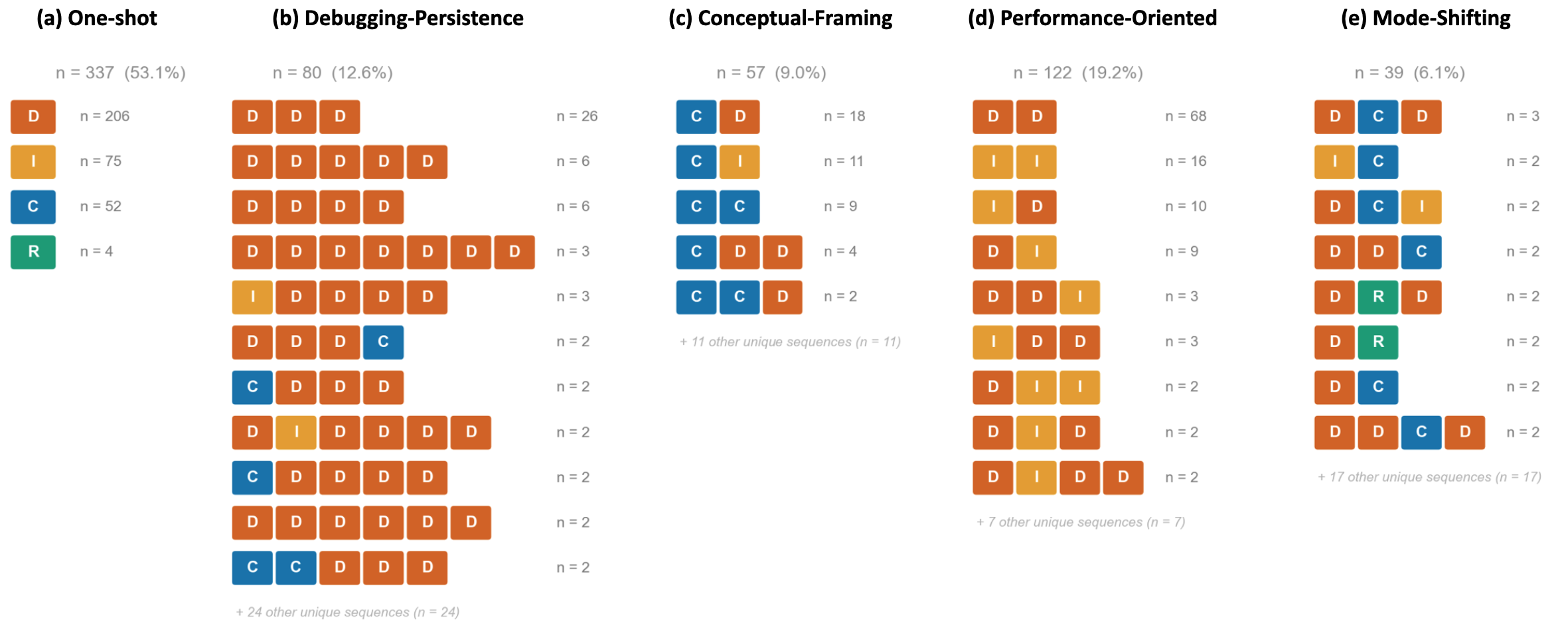}
\caption{Representative sequences within each attempt-level trajectory pattern. Each block indicates one prompt category in a task-specific session: Conceptual (C), Implementation (I), Debugging (D), and Reflective (R). The five panels show the taxonomy: (a) One-shot, (b) Debugging-Persistence, (c) Conceptual-Framing, (d) Performance-Oriented, and (e) Mode-Shifting.}
\label{fig:attempt_taxonomy}
\end{figure}

Figure~\ref{fig:attempt_taxonomy} shows representative sequences within each taxonomy pattern. For readability, only sequence variants appearing at least twice are shown, and lower-frequency sequences are summarized as other unique sequences. 

Overall, One-shot sessions were the most common pattern, accounting for 337 attempts (53.1\%). Within this group, single Debugging prompts were dominant (n = 206), followed by single Implementation prompts (n = 75), single Conceptual prompts (n = 52), and single Reflective prompts (n = 4). This indicates that more than half of task-specific AI use consisted of a single request, and that even these short interactions were primarily oriented toward immediate debugging or implementation needs.

Debugging-Persistence attempts formed the second pattern in the taxonomy and accounted for 80 attempts (12.6\%). This group was centered on repeated Debugging sequences, most notably D--D--D (n = 26), D--D--D--D--D (n = 6), and D--D--D--D (n = 6). Some Debugging-Persistence attempts began with Conceptual or Implementation prompts, such as C--D--D--D or I--D--D--D--D, but subsequently showed sustained persistence in debugging-oriented help-seeking.

Conceptual-Framing attempts accounted for 57 attempts (9.0\%). The most common variants were C--D (n = 18), C--I (n = 11), and C--C (n = 9), indicating that these sessions began with conceptual clarification, then moved into action-oriented support or remained within Conceptual help.

Among the remaining multi-prompt sessions, Performance-Oriented attempts formed the largest empirical group (n = 122, 19.2\%). The most frequent sequence variant in this group was D--D (n = 68), followed by I--I (n = 16), I--D (n = 10), and D--I (n = 9). These sequences suggest that many multi-turn interactions remained focused on Implementation and Debugging rather than moving toward Conceptual or Reflective help.

Mode-Shifting attempts were the least frequent pattern (n = 39, 6.1\%) and were more heterogeneous, including variants such as D--C--D, I--C, D--C--I, and D--R--D. These sequences indicate that only a small proportion of attempts involved broader shifts across help-seeking modes, particularly shifts from Implementation or Debugging into Conceptual or Reflective support.

\begin{figure}[t]
  \centering
  \includegraphics[width=1\linewidth]{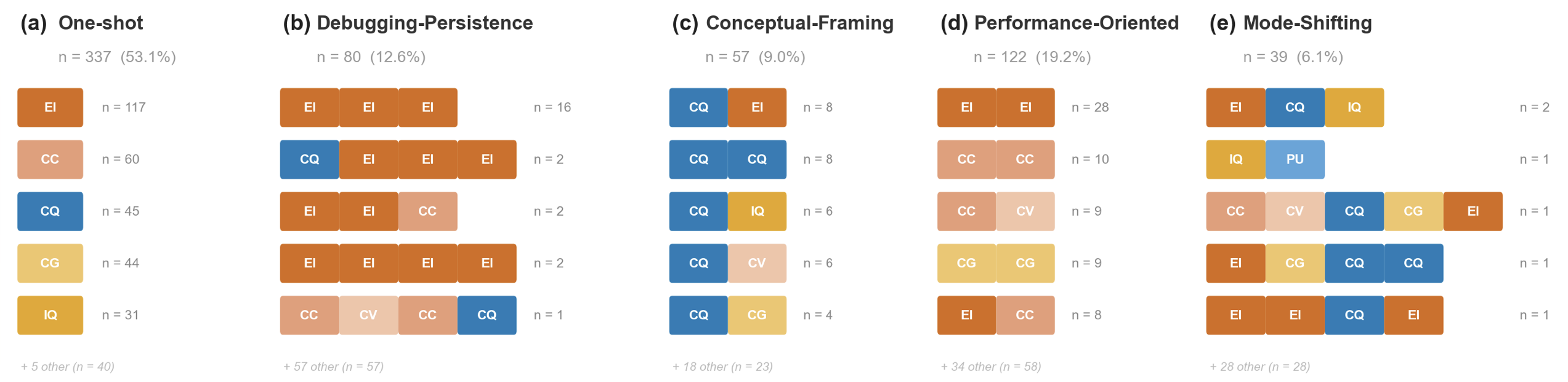}
\caption{Most frequent code-level sequences within each attempt-level trajectory pattern. Each panel shows the most common fine-grained prompt-code sequences observed within one taxonomy pattern: (a) One-shot, (b) Debugging-Persistence, (c) Conceptual-Framing, (d) Performance-Oriented, and (e) Mode-Shifting.}
\label{fig:code_taxonomy}
\end{figure}

Figure~\ref{fig:code_taxonomy} presents the five most frequent code-level sequences within each trajectory pattern. This figure complements the category-level taxonomy in Figure~\ref{fig:attempt_taxonomy} by showing which fine-grained prompt codes most commonly contributed to each broader pattern. Several points are noteworthy. First, One-shot sessions were not homogeneous. The most frequent single-code sequences included Error Interpretation (EI), Code Correction (CC), Concept Question (CQ), and Code Generation (CG), indicating that one-turn AI use served multiple functions, ranging from error diagnosis and code repair to conceptual clarification and direct solution construction. Second, Conceptual-Framing trajectories were mainly characterized by sequences that began with conceptual or problem-understanding prompts and then moved into implementation or debugging-related support. This suggests that conceptual help often functioned as an entry point into task work, rather than remaining isolated from subsequent programming activity. Third, Performance-Oriented sessions included sequences centered on Code Generation (CG) and Implementation Question (IQ), including repeated or combined implementation-oriented requests. These sequences suggest that this pattern often involved using AI to move directly toward solution construction. Fourth, Debugging-Persistence trajectories were composed of repeated or closely related debugging subtypes, especially sequences involving Error Interpretation (EI), Code Correction (CC), and Code Verification (CV). Finally, Mode-Shifting sequences were highly heterogeneous, with no single dominant code-level pathway. This heterogeneity suggests that category-level shifts could arise from multiple individualized routes across conceptual, implementation, debugging, and reflective forms of help-seeking.

Taken together, the category- and code-level sequence results suggest that attempt-level AI-assisted help-seeking was dominated by One-shot requests and Performance-Oriented multi-turn activity. Conceptual-Framing and Mode-Shifting were comparatively less common, whereas a substantial subset of multi-prompt attempts entered sustained debugging persistence.

\newcolumntype{L}{>{\raggedright\arraybackslash}X}
\begin{table}[t]
  \centering
  \caption{Attempt-level trajectory patterns and programming performance.}
  \label{tab:attempt-patterns}
  \footnotesize
  \renewcommand{\arraystretch}{1.12}
  \setlength{\tabcolsep}{4pt}
  \begin{tabularx}{\linewidth}{>{\raggedright\arraybackslash}p{0.22\linewidth}LLLL}
    \toprule
    \textbf{Pattern}
    & \textbf{$n$ (\%)}
    & \textbf{Prompts $M$ ($SD$)}
    & \textbf{Score $M$ ($SD$)}
    & \textbf{Submissions $M$ ($SD$)} \\
    \midrule

    \textbf{One-shot}
    & 337 (53.1)
    & 1.00 (0.00)
    & 4.34 (1.09)
    & 4.82 (4.16) \\

    \quad \emph{Debugging (D)}
    & \quad \emph{206 (32.4)}
    & \quad ---
    & \quad \emph{4.36 (1.09)}
    & \quad \emph{4.88 (3.45)} \\

    \quad \emph{Implementation (I)}
    & \quad \emph{75 (11.8)}
    & \quad ---
    & \quad \emph{4.28 (1.08)}
    & \quad \emph{4.22 (3.45)} \\

    \quad \emph{Conceptual (C)}
    & \quad \emph{52 (8.2)}
    & \quad ---
    & \quad \emph{4.31 (1.04)}
    & \quad \emph{5.52 (6.86)} \\

    \quad \emph{Reflective (R)}
    & \quad \emph{4 (0.6)}
    & \quad ---
    & \quad \emph{\textbf{4.50 (1.91)}}
    & \quad \emph{4.00 (1.83)} \\

    \addlinespace[2pt]

    \textbf{Debugging-Persistence}
    & 80 (12.6)
    & \textbf{5.10 (2.74)}
    & 4.49 (1.02)
    & \textbf{11.90 (12.04)} \\

    \textbf{Conceptual-Framing}
    & 57 (9.0)
    & 2.51 (0.97)
    & 4.41 (0.91)
    & 5.91 (5.88) \\

    \textbf{Performance-Oriented}
    & 122 (19.2)
    & 2.22 (0.57)
    & 4.39 (1.01)
    & 6.78 (6.15) \\

    \textbf{Mode-Shifting}
    & 39 (6.1)
    & 3.56 (1.64)
    & 4.42 (0.91)
    & 6.54 (5.55) \\

    \bottomrule
  \end{tabularx}
\end{table}

\subsubsection{Performance Across Taxonomy Patterns}

Because students could submit solutions multiple times in the automated assessment environment, we examined programming performance from two complementary perspectives: task score and the number of code submissions. The task score reflected the correctness of the submitted solution as evaluated by the automated assessment system, whereas the number of code submissions captured how many times students submitted code while working toward a solution. In addition, we report the number of prompts per session, which captures the extent of AI interaction within each attempt.

Table~\ref{tab:attempt-patterns} summarizes the number of task-specific student prompts, task scores, and the number of code submissions across the attempt-level trajectory patterns. Each programming task was scored out of 5 points. Although the highest mean score appeared in the One-shot Reflective subgroup ($M = 4.50$, $SD = 1.91$), this subgroup was very small ($n = 4$), and most groups clustered within a narrow range from $M = 4.28$ to $M = 4.49$. A Kruskal--Wallis test indicated that task score did not differ significantly across groups ($H = 2.16$, $p = .950$). This pattern is understandable in the automated assessment setting, where students could submit multiple times and eventually earn full or partial credit on most tasks.

The number of prompts per session varied substantially across patterns. One-shot sessions consisted of exactly one prompt by definition. Among multi-prompt patterns, Debugging-Persistence involved the most AI interaction ($M = 5.10$, $SD = 2.74$), followed by Mode-Shifting ($M = 3.56$, $SD = 1.64$) and Conceptual-Framing ($M = 2.51$, $SD = 0.97$). Performance-Oriented sessions, despite being the largest multi-prompt group, were relatively short ($M = 2.22$, $SD = 0.57$), suggesting that these interactions typically involved only brief implementation--debugging exchanges.

The number of code submissions also differed significantly across groups ($H = 82.49$, $p < .001$). Debugging-Persistence required the most code submissions ($M = 11.90$, $SD = 12.04$), substantially more than the overall One-shot pattern ($M = 4.82$, $SD = 4.16$). Performance-Oriented ($M = 6.78$, $SD = 6.15$) and Mode-Shifting attempts ($M = 6.54$, $SD = 5.55$) required a moderate number of code submissions, whereas Conceptual-Framing attempts ($M = 5.91$, $SD = 5.88$) were closer to One-shot. Notably, Debugging-Persistence stood out on both dimensions: it involved the most prompts and the most code submissions, suggesting that sustained debugging-oriented help-seeking was associated with higher process costs across both AI interaction and code revision.

Because the number of code submissions may also be influenced by task difficulty, we conducted a robustness check by comparing easier and harder tasks. Task difficulty was determined by the course instructor based on the expected conceptual and implementation complexity of each programming task. Trajectory pattern was not significantly associated with task difficulty ($\chi^2 = 3.84$, $df = 4$, $p = .428$, Cramer's $V = 0.078$). In addition, Debugging-Persistence attempts required more submissions than One-shot attempts within both harder tasks ($p = .008$) and easier tasks ($p < .001$). These results suggest that the higher number of code submissions associated with Debugging-Persistence was not solely attributable to task difficulty.

\subsection{SRL-Informed Case Interpretation}

To illustrate the educational significance of the attempt-level trajectory patterns from an SRL-informed perspective, we present five illustrative cases, one for each pattern: One-shot, Debugging-Persistence, Conceptual-Framing, Performance-Oriented, and Mode-Shifting. We do not treat individual prompts as direct evidence of students' internal SRL states. Instead, we use the SRL-informed lens introduced in Figure~\ref{fig:framework} to interpret observable help-seeking structures. The cases therefore illustrate how different observable trajectories may align with distinct regulatory opportunities or risks.

The student prompts and AI responses in the illustrative cases were originally in Japanese and were translated into English by the authors for presentation. Because the original AI responses were often lengthy, only relevant excerpts are shown, with ellipses indicating omitted portions.

\begin{table*}[t]
  \centering
  \caption{Illustrative case of a One-shot trajectory for a numeric input task.}
  \label{tab:case_one_shot_alternative}
  \footnotesize
  \renewcommand{\arraystretch}{1.18}
  \begin{tabular*}{\textwidth}{@{\extracolsep{\fill}}p{0.16\textwidth}p{0.78\textwidth}}
    \toprule
    \textbf{Step} & \textbf{Dialogue and submission record} \\
    \midrule

    \textbf{Task}
    &
    Write a program that repeatedly receives integer input until \texttt{0} is entered, then displays the maximum value, minimum value, and number of entered integers.
    \\

    \midrule

    \textbf{Turn 1: CO}
    &
    \textbf{Student:}
    
    ``\texttt{numbers = []}

    \texttt{count = 0}

    \texttt{while True:}

    \texttt{\quad num = int(input("Input a number: "))}

    \texttt{\quad if num == 0:}

    \texttt{\quad\quad break}

    \texttt{\quad numbers.append(num)}

    \texttt{\quad count += 1}

    \texttt{...}

    Are there other ways to write this?''

    \par\smallskip
    \textbf{AI:}
    ``Yes, there are several other ways to write the same functionality. Below are several Python code styles with the same behavior. \ldots''
    \\
    \midrule
    \textbf{Submission record}
    &
    \textbf{Submission 1: Passed.}
    \\

    \bottomrule
  \end{tabular*}
\end{table*}

\subsubsection{A One-shot Trajectory}

Table~\ref{tab:case_one_shot_alternative} presents a One-shot trajectory. In this case, the student had already written a solution for a numeric input task and used a single prompt to ask whether there were other ways to write the program. Unlike a one-shot request for code generation or error correction, this prompt was designed to explore alternative implementations after an initial solution had been constructed.

From an SRL-informed perspective, this case illustrates that One-shot help-seeking is not necessarily shallow or purely reactive. Although the session did not involve turn-by-turn development, the student used AI support to compare possible solution styles and broaden their understanding of implementation alternatives. Thus, although many One-shot trajectories involved brief debugging-oriented help that supported immediate task completion, this case shows that a single-prompt interaction can also reflect more reflective or optimization-oriented help-seeking.

\begin{table*}[t]
  \centering
  \caption{Illustrative case of a Debugging-Persistence trajectory for a string-output task.}
  \label{tab:case_repair_lock_repeat_input}
  \footnotesize
  \renewcommand{\arraystretch}{1.18}
  \begin{tabular*}{\textwidth}{@{\extracolsep{\fill}}p{0.16\textwidth}p{0.78\textwidth}}
    \toprule
    \textbf{Step} & \textbf{Dialogue and submission record} \\
    \midrule

    \textbf{Task}
    &
    Define a function \texttt{repeat\_input(s)} that prints the input string in the required format:
    \texttt{Your input is "}\emph{content of s}\texttt{"}.
    \\

    \midrule

    \textbf{Turn 1: CV}
    &
    \textbf{Student:}
    
    ``\texttt{def repeat\_input(s):}

    \texttt{\quad print("Your input is \{\}".format(s))}''

    \par\smallskip
    \textbf{AI:}
    ``Your \texttt{repeat\_input} function receives the argument \texttt{s} and formats its content for output. \ldots''
    \\

    \midrule

    \textbf{Turn 2: CC}
    &
    \textbf{Student:}
    
    ``\texttt{def repeat\_input(s):}

    \texttt{\quad print("Your input is "\{\}" ".format(s))}''

    \par\smallskip
    \textbf{AI:}
    ``Your function needs a small correction. The quotation marks in the string-formatting part are not closed correctly. The correct code is: \texttt{print("Your input is \{\}".format(s))} \ldots''
    \\

    \midrule

    \textbf{Turn 3: CC}
    &
    \textbf{Student:}
    
    ``\texttt{def repeat\_input(s):}

    \texttt{\quad print("Your input is ”\{\}” ".format(s))}''

    \par\smallskip
    \textbf{AI:}
    ``The quotation marks in your function are not valid. In Python, standard single quotes or double quotes should be used to define strings. \ldots''
    \\

    \midrule

    \textbf{Turn 4: IQ}
    &
    \textbf{Student:}
    ``I want to display \texttt{Your input is "}content of string \texttt{s}\texttt{"}.''

    \par\smallskip
    \textbf{AI:}
    ``To display the content of \texttt{s} enclosed in quotation marks, you can write the function as follows:
    \texttt{print('Your input is "\{\}"'.format(s))}. \ldots''
    \\

    \midrule

    \textbf{Turn 5: CV}
    &
    \textbf{Student:}
    
    ``\texttt{def repeat\_input(s):}

    \texttt{\quad print('Your input is " \{\} " '.format(s))}''

    \par\smallskip
    \textbf{AI:}
    ``Your code is almost correct, but you need to be careful about the placement of quotation marks and spaces. \ldots''
    \\

    \midrule

    \textbf{Turn 6: CV}
    &
    \textbf{Student:}
    
    ``\texttt{def repeat\_input(s):}

    \texttt{\quad print('Your input is "\{\}" '.format(s))}''

    \par\smallskip
    \textbf{AI:}
    ``Your code is almost correct, but the output contains an extra space. \ldots''
    \\

    \midrule

    \textbf{Turn 7: CV}
    &
    \textbf{Student:}
    
    ``\texttt{def repeat\_input(s):}

    \texttt{\quad print('Your input is "\{\}"'.format(s))}''

    \par\smallskip
    \textbf{AI:}
    ``Your code is correct and displays the argument \texttt{s} enclosed in quotation marks. \ldots''
    \\

    \midrule

    \textbf{Submission record}
    &
    \textbf{Submissions 1--6: Failed.}

    \par\smallskip
    \textbf{Submission 7: Passed.}
    \\

    \bottomrule
  \end{tabular*}
\end{table*}

\subsubsection{A Debugging-Persistence Trajectory}

Table~\ref{tab:case_repair_lock_repeat_input} presents a Debugging-Persistence trajectory. In this case, the student worked on a short string-output task that required printing the input string in a specific format. Although the task was conceptually simple, the interaction was dominated by repeated surface-level corrections to quotation marks, spacing, and output formatting. The student repeatedly modified minor surface features of the code and returned to the AI for correction or verification, rather than seeking a broader explanation of how nested quotation marks and string formatting work.

This case illustrates how Debugging-Persistence can occur even when the learner is close to the correct solution. The interaction remained focused on repeated small modifications and external verification of formatting-related issues, with little observable shift toward alternative debugging strategies or conceptual reframing. 

\begin{table*}[t]
  \centering
  \caption{Illustrative case of a Conceptual-Framing trajectory for a string-containment task.}
  \label{tab:case_conceptual_framing_string}
  \footnotesize
  \renewcommand{\arraystretch}{1.18}
  \begin{tabular*}{\textwidth}{@{\extracolsep{\fill}}p{0.22\textwidth}p{0.72\textwidth}}
    \toprule
    \textbf{Step} & \textbf{Dialogue and submission record} \\
    \midrule

    \textbf{Task}
    &
    Write \texttt{print\_is\_contained(sentence)} that prints ``yes'' if \texttt{"a"} appears in the input string, and ``no'' otherwise.
    \\

    \midrule

    \textbf{Turn 1: CQ}
    &
    \textbf{Student:}
    ``How do I receive a string as input?''

    \par\smallskip
    \textbf{AI:}
    ``To receive a string as input in Python, use the \texttt{input()} function. This function waits for user input and returns the entered string. \ldots''
    \\

    \midrule

    \textbf{Turn 2: CV}
    &
    \textbf{Student:}
    
``\texttt{def print\_is\_contained(sentence):}

    \texttt{\quad if "a" in sentence:}

    \texttt{\quad\quad print("yes")}

    \texttt{\quad else:}

    \texttt{\quad\quad print("no")}

    Is this correct?''

    \par\smallskip
    \textbf{AI:}
    ``Your function \texttt{print\_is\_contained} is written with correct syntax to check whether the character \texttt{"a"} is contained in the given string. It prints \texttt{"yes"} if \texttt{"a"} is found and \texttt{"no"} otherwise. \ldots''
    \\
    \midrule
    \textbf{Submission record}
    &
    \textbf{Submission 1: Passed.}
    \\
    \bottomrule
  \end{tabular*}
\end{table*}

\subsubsection{A Conceptual-Framing Trajectory: Conceptual Clarification Before Code Verification}

Table~\ref{tab:case_conceptual_framing_string} presents a Conceptual-Framing trajectory. In this case, the student first asked how to receive a string as input, clarifying a task-relevant programming concept before submitting code. The subsequent prompt did not request a solution. Instead, the student shared a self-constructed function and asked whether it was correct. The submitted code then passed on the first attempt.

This case illustrates a trajectory from conceptual clarification to performance-oriented checking. The student used AI support to clarify a prerequisite for the task and then to verify an implementation that they had produced. It reflects a short but structured help-seeking trajectory in which task-relevant understanding was established before solution verification.

\begin{table*}[t]
  \centering
  \caption{Illustrative case of a Performance-Oriented trajectory for a numeric input task.}
  \label{tab:case_performance_oriented_numeric}
  \footnotesize
  \renewcommand{\arraystretch}{1.18}
  \begin{tabular*}{\textwidth}{@{\extracolsep{\fill}}p{0.16\textwidth}p{0.78\textwidth}}
    \toprule
    \textbf{Step} & \textbf{Dialogue and submission record} \\
    \midrule

    \textbf{Task}
    &
    Write a program that asks the user to enter integers from the keyboard, then displays the maximum value, the minimum value, and the number of integers entered. The program should terminate when \texttt{0} is entered, and the output should match the following example:

    \par\smallskip
    \texttt{Input a number: 1}

    \texttt{Input a number: 2}

    \texttt{Input a number: 3}

    \texttt{Input a number: 0}

    \texttt{max: 3}

    \texttt{min: 1}

    \texttt{\# of inputs: 3}
    \\

    \midrule

    \textbf{Turn 1: CG}
    &
    \textbf{Student:}
    ``Write a program that takes integer input from the keyboard and finds and displays the maximum value, minimum value, and the number of integers entered.''

    \par\smallskip
    \textbf{AI:}
    ``Here is a simple Python program that keeps reading integers and then displays the maximum value, minimum value, and the number of inputs. \ldots''
    \\

    \midrule

    \textbf{Turn 2: CG}
    &
    \textbf{Student:}
    ``However, the program should terminate when \texttt{0} is entered. Also, please match the required output format.''

    \par\smallskip
    \textbf{AI:}
    ``Understood. The program should stop when \texttt{0} is entered and should print the output in the required format. \ldots''
    \\

    \midrule

    \textbf{Turn 3: EI}
    &
    \textbf{Student:}
    ``The output shows \texttt{Input a number:} with a newline instead of \texttt{Input a number: } with a trailing space.''

    \par\smallskip
    \textbf{AI:}
    ``Sorry, the output format was slightly different from what was required. Please use \texttt{input("Input a number: ")} so that the input appears after the prompt on the same line. \ldots''
    \\

    \midrule

    \textbf{Submission record}
    &
    \textbf{Submission 1--2: Failed.} 
    \par\smallskip
    \textbf{Submission 3: Passed.} 
    \\

    \bottomrule
  \end{tabular*}
\end{table*}

\subsubsection{A Performance-Oriented Trajectory}

Table~\ref{tab:case_performance_oriented_numeric} presents a Performance-Oriented trajectory. In this case, the student moved directly into solution construction by asking the AI to write a program, rather than first presenting an initial code attempt or asking for conceptual clarification. The first prompt provided only part of the task requirements, and the student subsequently added the termination condition and output-format constraints. The final error-related prompt focused on correcting a formatting mismatch in the generated solution.

The student reached a passing solution after three submissions. However, the interaction mainly involved generating, refining, and aligning AI-produced code with the grader's expected output, with limited evidence that the student articulated or monitored the underlying solution strategy. Thus, the case reflects task-completion efficiency, while leaving open the possibility that some problem-solving steps were offloaded to the AI.

\subsubsection{A Mode-Shifting Trajectory: Moving from Verification to Conceptual Clarification}

Table~\ref{tab:case_mode_shifting_string} presents a Mode-Shifting trajectory. The interaction began with a debugging-oriented verification request, in which the student asked about a function call to \texttt{print\_is\_contained}. After the AI provided an example using \texttt{input\_str} as the parameter name, the student did not simply copy the suggested code. Instead, the student shifted to a conceptual question about whether the parameter name had to be \texttt{input\_str} or whether other names could be used. Importantly, the final prompt used \texttt{word} as the parameter name, showing that the conceptual clarification was incorporated into the student's own code. This pattern suggests a more flexible form of help-seeking than that provided by debugging alone. The student paused the debugging-oriented interaction to clarify a transferable programming concept, then returned to the task with a modified solution. 

Taken together, the five cases clarify the SRL-informed meaning of the quantitative trajectory patterns. They suggest that the educational significance of AI support depends not only on whether students use AI, but also on whether their help-seeking trajectories support planning, monitoring, and reflection or remain centered on reactive troubleshooting and repeated correction.

\begin{table*}[t]
  \centering
  \caption{Illustrative case of a Mode-Shifting trajectory for a string-containment task.}
  \label{tab:case_mode_shifting_string}
  \footnotesize
  \renewcommand{\arraystretch}{1.18}
  \begin{tabular*}{\textwidth}{@{\extracolsep{\fill}}p{0.16\textwidth}p{0.78\textwidth}}
    \toprule
    \textbf{Step} & \textbf{Dialogue and submission record} \\
    \midrule

    \textbf{Task}
    &
    Define a function \texttt{print\_is\_contained(sentence)} that prints ``yes'' if the input string contains \texttt{"a"}, and ``no'' otherwise.
    \\

    \midrule

    \textbf{Turn 1: CV}
    &
    \textbf{Student:}
    ``\texttt{print\_is\_contained("aiueo")}''

    \par\smallskip
    \textbf{AI:}
    ``This code assumes that a function named \texttt{print\_is\_contained} has already been defined. The function checks whether the given string contains the character \texttt{'a'} and prints the result. Since \texttt{"aiueo"} contains \texttt{'a'}, it should print \texttt{"yes"}. Here is an example \ldots''
    \\

    \midrule

    \textbf{Turn 2: CQ}
    &
    \textbf{Student:}
    ``I understand! Does the name inside the parentheses have to be \texttt{input\_str}? Are there other ways to write it?''

    \par\smallskip
    \textbf{AI:}
    ``No, the function argument name does not have to be \texttt{input\_str}. The argument name can be freely chosen by the programmer. \texttt{input\_str} is only a common example. Other possible names include \texttt{text}, \texttt{sentence}, \texttt{word}, and \texttt{phrase}. \ldots''
    \\

    \midrule

    \textbf{Turn 3: CV}
    &
    \textbf{Student:}
    
    ``\texttt{def print\_is\_contained(word):}

    \texttt{\quad if 'a' in word:}

    \texttt{\quad\quad print("yes")}

    \texttt{\quad else:}

    \texttt{\quad\quad print("no")}

    Is this also correct?''

    \par\smallskip
    \textbf{AI:}
    ``Yes, the provided code is also correct. You changed the argument name to \texttt{word} and modified the code consistently. \ldots''
    \\

    \midrule

    \textbf{Submission record}
    &
    \textbf{Submissions 1--2: Failed.}

    \par\smallskip
    \textbf{Submission 3: Passed.}
    \\

    \bottomrule
  \end{tabular*}
\end{table*}

\section{Discussion}

This study examined AI-assisted help-seeking trajectories in introductory programming from an SRL-informed perspective. We analyzed how students' prompts unfolded across turns and attempts, and how different interaction structures were associated with task scores and the number of code submissions. Overall, the results show that the educational significance of AI support lies not only in whether students use AI or how often they do, but also in how their help-seeking trajectories develop during programming attempts.

\subsection{RQ1: How Do AI-Assisted Help-Seeking Trajectories Unfold?}

For RQ1, the results show that students' AI-assisted help-seeking was strongly concentrated around immediate problem resolution. Debugging was the dominant help-seeking category, whereas Reflective help was rare. This distribution suggests that students primarily used AI as a reactive programming aid when they encountered errors, incorrect outputs, or uncertainty about whether their code was correct.

More importantly, the sequential analyses showed that these help-seeking behaviors were not merely isolated requests. The Sankey diagram, transition network, and lag sequential analysis all suggested that students' interactions were not characterized by flexible movement among conceptual understanding, implementation, debugging, and reflection. Instead, many trajectories were centered on reactive troubleshooting, especially repeated debugging-oriented support. This pattern suggests that students often use AI to obtain immediate fixes for coding problems, rather than to support a broader regulatory process of error diagnosis, progress monitoring, and reflection on possible solutions.

This finding refines how AI use in programming education should be understood. The issue is not simply that students ask many debugging questions, because debugging is a normal and necessary part of learning to program. The more important concern is that AI-assisted debugging may become a reactive troubleshooting cycle, in which students repeatedly rely on AI for corrections without sufficiently diagnosing the problem, monitoring their progress, or confirming why the revised code works. From an SRL-informed perspective, the educational risk lies in help-seeking trajectories that may reduce students' opportunities to diagnose problems, verify changes, and regulate their own problem-solving.

\subsection{RQ2: Trajectory Patterns and Programming Performance}

For RQ2, we summarized task-specific AI interactions into five attempt-level trajectory patterns: One-shot, Debugging-Persistence, Conceptual-Framing, Performance-Oriented, and Mode-Shifting. This taxonomy shows that students' AI-assisted help-seeking was not organized in a single uniform way. 

In terms of distribution, One-shot sessions were the most common form of task-specific AI use, accounting for more than half of the attempts. Among multi-prompt sessions, Performance-Oriented attempts formed the largest group, suggesting that many extended interactions remained centered on implementation and debugging. Debugging-Persistence was less frequent but represented a distinct trajectory in which students engaged in repeated debugging-oriented repair.

These trajectory patterns were associated more clearly with the number of code submissions than with the task score. Task scores were similarly high across groups, which is understandable in an assessment environment where students could submit multiple times and eventually reach a correct or partially correct solution. In contrast, the number of code submissions differed substantially across patterns. Debugging-Persistence attempts required the most submissions, whereas One-shot attempts required the fewest. This suggests that the main difference among trajectory patterns lies less in the assessed outcome and more in the process cost of reaching that outcome.

This distinction is important for interpreting programming performance in AI-supported learning environments. A successful final score does not necessarily mean that the problem-solving process was efficient or well-regulated. What matters educationally is not only whether students can complete a task with AI support, but how they learn through that interaction: whether they clarify concepts, monitor errors, evaluate suggestions, and gradually take ownership of the solution.

\subsection{RQ3: How Can These Trajectories Be Interpreted from an SRL-Informed Perspective?}

For RQ3, the illustrative cases show that similar task outcomes can arise from different regulatory structures. For example, a low number of code submissions was observed in both the Conceptual-Framing and Performance-Oriented trajectories, but these patterns carried distinct meanings. In the Conceptual-Framing case, the student first clarified a task-relevant concept and then verified their own implementation. In the Performance-Oriented case, the student moved more directly into AI-supported solution construction, leaving open the possibility that some problem-solving steps were offloaded to the AI. In the Debugging-Persistence case, the student was close to the required output but remained in a narrow trial-and-repair loop around formatting details. By contrast, the Mode-Shifting case showed a student moving from verification to conceptual clarification and then back to verification, suggesting that shifts across help-seeking modes can sometimes support more flexible regulation. The One-shot case further indicates that even a single prompt can be reflective or optimization-oriented when students use AI to explore alternative approaches.

Taken together, these cases suggest that the educational value of AI support depends on how students engage with AI during problem solving, not only on whether they complete the task. The number of interactions with AI provides useful evidence about the extent of AI use, but it does not, by itself, reveal whether students used AI to clarify concepts, test ideas, verify understanding, or repeatedly seek corrections. An SRL-informed lens helps distinguish these possibilities by shifting attention from the amount of AI use to the structure of students' help-seeking trajectories.

\subsection{Implications for AI-Supported Programming Education}

The findings suggest several implications for the design and use of AI tools in programming education. First, AI-supported programming systems should attend to help-seeking trajectories rather than rely solely on individual prompts. The results show that students' AI use often unfolded in recognizable patterns, such as One-shot, Conceptual-Framing, Performance-Oriented, Debugging-Persistence, and Mode-Shifting. This suggests that a student's current prompt should be interpreted within the broader interaction sequence. 

Second, AI systems should be designed to detect and respond to repair-oriented loops. Debugging is a normal and necessary part of programming learning, and debugging-oriented AI use should not be treated as inherently problematic. However, the Debugging-Persistence pattern suggests that repeated local repair can become costly when students continue to adjust small surface features of the code without consolidating the underlying rule or requirement. In such cases, an AI assistant could shift from giving another correction to supporting reflection or conceptual clarification. For example, after repeated code-correction or verification requests, the system might summarize the recurring issue, ask the student to explain the relevant rule, or prompt them to compare the expected and actual outputs before providing another fix \cite{li2025coderunner,ma2025scaffolding}. Such interventions could help students move from trial-and-error toward more deliberate monitoring of their own code.

Third, AI tools should balance efficient task support with opportunities for students to explain their work and take ownership of it. The Performance-Oriented case shows that direct code generation and output alignment can help students complete tasks efficiently, but this efficiency may also leave open the possibility that some problem-solving steps are offloaded to the AI. Rather than simply prohibiting code generation, educational AI tools could embed lightweight scaffolds after providing code, such as asking students to identify which lines handle input, loop termination, conditional logic, or output formatting \cite{ma2026design}. Similarly, when students ask for alternative implementations, as in the One-shot case, the system could encourage comparison across approaches, helping students understand why one solution is simpler, more readable, or more general than another.

Finally, instructors may benefit from trajectory-level analytics rather than raw AI usage counts. A student who asks one reflective question about alternative implementations differs from a student who repeatedly cycles through code correction prompts, even if both are counted as AI users. Similarly, a student who begins with conceptual clarification may require different instructional support from one who immediately requests a complete solution. Dashboards or reports that summarize help-seeking trajectories could provide instructors with more actionable information than simple prompt frequencies \cite{chen2024stugptviz}. 

\subsection{Limitations and Future Work}

Several limitations should be considered. First, this study was conducted in a single introductory Python course at one university. Future studies should examine these patterns across different courses, programming languages, and institutions. 

In addition, we used task scores and the number of code submissions to examine students' programming performance. However, these measures do not directly capture conceptual understanding, retention, transfer, or long-term programming ability. Future work should link AI-assisted help-seeking trajectories to richer learning outcomes.

The analysis was also limited to interactions captured through the logged AI interface. Students may have used other resources, including classmates, instructors, lecture materials, web searches, or off-platform AI tools. These unobserved sources may have shaped both the observed help-seeking trajectories and the corresponding submission records. Future research could combine AI logs with classroom observations, interviews, surveys, or LMS traces to better understand how AI fits into students' broader help-seeking ecology.

Another limitation concerns the scope of the coding. Our analysis focused on the functional role of student prompts rather than the pedagogical quality of AI responses. The illustrative cases suggest that AI responses can influence how students continue the interaction: some responses may support clarification or verification, whereas others may encourage direct adoption, repeated local repair, or misleading changes. Future work should jointly analyze student prompts and AI responses, including the directness, accuracy, scaffolding level, and explanatory quality of AI feedback. Such analysis would clarify not only what students ask for, but also how AI responses shape subsequent regulation.

The SRL-informed interpretation and attempt-level taxonomy should be understood as trace-based analytical lenses rather than direct measures of students' internal regulatory states. The five trajectory patterns provide an interpretable way to connect prompt-level behavior with submission processes, while recognizing that students' planning, monitoring, and reflection may also occur outside the logged AI interaction. Future work could further validate these patterns using interviews, think-aloud data, self-explanations, or scaffolding interventions.

\section{Conclusion}

This study examined AI-assisted help-seeking trajectories in introductory programming from an SRL-informed perspective. By linking task-specific student prompts with code submissions, we identified five recurring trajectory patterns: One-shot, Debugging-Persistence, Conceptual-Framing, Performance-Oriented, and Mode-Shifting. The results showed that students' AI use was largely oriented toward debugging and immediate problem resolution, and that trajectory patterns were more clearly associated with the number of code submissions than with task scores. These findings suggest that the educational significance of AI support cannot be understood only by asking whether students use AI or whether they eventually complete a task. Rather, it depends on how students engage with AI during problem-solving. Supporting students in moving beyond repeated debugging requests toward conceptual clarification, self-explanation, and verification may therefore be important for making AI-assisted programming support more compatible with self-regulated learning.

\backmatter

%\bmhead{Acknowledgements}

\section*{Declarations}

\begin{itemize}
\item Competing interests

The authors declare no competing interests.

\item Ethics approval and consent to participate

This study was approved by the Institutional Review Board at the authors’ institution (institution name blinded for peer review). Informed consent was obtained from all participants prior to their participation in the study. 

\item Data availability
The data and materials supporting the findings of this study are not publicly available due to privacy and ethical restrictions related to human participant data.

\end{itemize}

\bigskip

%\begin{appendices}
%
%\section{Section title of first appendix}\label{secA1}




%\end{appendices}


\bibliography{sn-bibliography}% common bib file


\begin{thebibliography}{73}
\ifx \bisbn   \undefined \def \bisbn  #1{ISBN #1}\fi
\ifx \binits  \undefined \def \binits#1{#1}\fi
\ifx \bauthor  \undefined \def \bauthor#1{#1}\fi
\ifx \batitle  \undefined \def \batitle#1{#1}\fi
\ifx \bjtitle  \undefined \def \bjtitle#1{#1}\fi
\ifx \bvolume  \undefined \def \bvolume#1{\textbf{#1}}\fi
\ifx \byear  \undefined \def \byear#1{#1}\fi
\ifx \bissue  \undefined \def \bissue#1{#1}\fi
\ifx \bfpage  \undefined \def \bfpage#1{#1}\fi
\ifx \blpage  \undefined \def \blpage #1{#1}\fi
\ifx \burl  \undefined \def \burl#1{\textsf{#1}}\fi
\ifx \doiurl  \undefined \def \doiurl#1{\url{https://doi.org/#1}}\fi
\ifx \betal  \undefined \def \betal{\textit{et al.}}\fi
\ifx \binstitute  \undefined \def \binstitute#1{#1}\fi
\ifx \binstitutionaled  \undefined \def \binstitutionaled#1{#1}\fi
\ifx \bctitle  \undefined \def \bctitle#1{#1}\fi
\ifx \beditor  \undefined \def \beditor#1{#1}\fi
\ifx \bpublisher  \undefined \def \bpublisher#1{#1}\fi
\ifx \bbtitle  \undefined \def \bbtitle#1{#1}\fi
\ifx \bedition  \undefined \def \bedition#1{#1}\fi
\ifx \bseriesno  \undefined \def \bseriesno#1{#1}\fi
\ifx \blocation  \undefined \def \blocation#1{#1}\fi
\ifx \bsertitle  \undefined \def \bsertitle#1{#1}\fi
\ifx \bsnm \undefined \def \bsnm#1{#1}\fi
\ifx \bsuffix \undefined \def \bsuffix#1{#1}\fi
\ifx \bparticle \undefined \def \bparticle#1{#1}\fi
\ifx \barticle \undefined \def \barticle#1{#1}\fi
\bibcommenthead
\ifx \bconfdate \undefined \def \bconfdate #1{#1}\fi
\ifx \botherref \undefined \def \botherref #1{#1}\fi
\ifx \url \undefined \def \url#1{\textsf{#1}}\fi
\ifx \bchapter \undefined \def \bchapter#1{#1}\fi
\ifx \bbook \undefined \def \bbook#1{#1}\fi
\ifx \bcomment \undefined \def \bcomment#1{#1}\fi
\ifx \oauthor \undefined \def \oauthor#1{#1}\fi
\ifx \citeauthoryear \undefined \def \citeauthoryear#1{#1}\fi
\ifx \endbibitem  \undefined \def \endbibitem {}\fi
\ifx \bconflocation  \undefined \def \bconflocation#1{#1}\fi
\ifx \arxivurl  \undefined \def \arxivurl#1{\textsf{#1}}\fi
\csname PreBibitemsHook\endcsname

%%% 1
\bibitem[\protect\citeauthoryear{Smith et~al.}{2017a}]{smith2017office}
\begin{barticle}
\bauthor{\bsnm{Smith}, \binits{M.}},
\bauthor{\bsnm{Chen}, \binits{Y.}},
\bauthor{\bsnm{Berndtson}, \binits{R.}},
\bauthor{\bsnm{Burson}, \binits{K.M.}},
\bauthor{\bsnm{Griffin}, \binits{W.}}:
\batitle{"office hours are kind of weird": Reclaiming a resource to foster
  student-faculty interaction.}
\bjtitle{InSight: A Journal of Scholarly Teaching}
\bvolume{12},
\bfpage{14}--\blpage{29}
(\byear{2017})
\end{barticle}
\endbibitem

%%% 2
\bibitem[\protect\citeauthoryear{Smith et~al.}{2017b}]{smith2017my}
\begin{bchapter}
\bauthor{\bsnm{Smith}, \binits{A.J.}},
\bauthor{\bsnm{Boyer}, \binits{K.E.}},
\bauthor{\bsnm{Forbes}, \binits{J.}},
\bauthor{\bsnm{Heckman}, \binits{S.}},
\bauthor{\bsnm{Mayer-Patel}, \binits{K.}}:
\bctitle{My digital hand: A tool for scaling up one-to-one peer teaching in
  support of computer science learning}.
In: \bbtitle{Proceedings of the 2017 ACM SIGCSE Technical Symposium on Computer
  Science Education}.
\bsertitle{SIGCSE '17},
pp. \bfpage{549}--\blpage{554}.
\bpublisher{Association for Computing Machinery},
\blocation{New York, NY, USA}
(\byear{2017}).
\doiurl{10.1145/3017680.3017800} .
\burl{https://doi.org/10.1145/3017680.3017800}
\end{bchapter}
\endbibitem

%%% 3
\bibitem[\protect\citeauthoryear{National Academies~of Sciences
  et~al.}{2018}]{national2018assessing}
\begin{bbook}
\bauthor{\bsnm{Sciences}, \binits{E.}},
\bauthor{\bsnm{Medicine}}, \betal:
\bbtitle{Assessing and Responding to the Growth of Computer Science
  Undergraduate Enrollments}.
\bpublisher{National Academies Press},
\blocation{Washington, DC}
(\byear{2018})
\end{bbook}
\endbibitem

%%% 4
\bibitem[\protect\citeauthoryear{Qian and Lehman}{2017}]{qian2017students}
\begin{barticle}
\bauthor{\bsnm{Qian}, \binits{Y.}},
\bauthor{\bsnm{Lehman}, \binits{J.}}:
\batitle{Students’ misconceptions and other difficulties in introductory
  programming: A literature review}.
\bjtitle{ACM Transactions on Computing Education (TOCE)}
\bvolume{18}(\bissue{1}),
\bfpage{1}--\blpage{24}
(\byear{2017})
\end{barticle}
\endbibitem

%%% 5
\bibitem[\protect\citeauthoryear{Tian et~al.}{2023}]{tian2304chatgpt}
\begin{botherref}
\oauthor{\bsnm{Tian}, \binits{H.}},
\oauthor{\bsnm{Lu}, \binits{W.}},
\oauthor{\bsnm{Li}, \binits{T.}},
\oauthor{\bsnm{Tang}, \binits{X.}},
\oauthor{\bsnm{Cheung}, \binits{S.}},
\oauthor{\bsnm{Klein}, \binits{J.}},
\oauthor{\bsnm{Bissyand{\'e}}, \binits{T.}}:
Is chatgpt the ultimate programming assistant—how far is it?
arXiv preprint arXiv:2304.11938
(2023)
\end{botherref}
\endbibitem

%%% 6
\bibitem[\protect\citeauthoryear{Yilmaz and Yilmaz}{2023}]{yilmaz2023augmented}
\begin{barticle}
\bauthor{\bsnm{Yilmaz}, \binits{R.}},
\bauthor{\bsnm{Yilmaz}, \binits{F.G.K.}}:
\batitle{Augmented intelligence in programming learning: Examining student
  views on the use of chatgpt for programming learning}.
\bjtitle{Computers in Human Behavior: Artificial Humans}
\bvolume{1}(\bissue{2}),
\bfpage{100005}
(\byear{2023})
\end{barticle}
\endbibitem

%%% 7
\bibitem[\protect\citeauthoryear{Kazemitabaar
  et~al.}{2024}]{kazemitabaar2024codeaid}
\begin{bchapter}
\bauthor{\bsnm{Kazemitabaar}, \binits{M.}},
\bauthor{\bsnm{Ye}, \binits{R.}},
\bauthor{\bsnm{Wang}, \binits{X.}},
\bauthor{\bsnm{Henley}, \binits{A.Z.}},
\bauthor{\bsnm{Denny}, \binits{P.}},
\bauthor{\bsnm{Craig}, \binits{M.}},
\bauthor{\bsnm{Grossman}, \binits{T.}}:
\bctitle{Codeaid: Evaluating a classroom deployment of an llm-based programming
  assistant that balances student and educator needs}.
In: \bbtitle{Proceedings of the 2024 Chi Conference on Human Factors in
  Computing Systems},
pp. \bfpage{1}--\blpage{20}
(\byear{2024})
\end{bchapter}
\endbibitem

%%% 8
\bibitem[\protect\citeauthoryear{Sun et~al.}{2024}]{sun2024would}
\begin{barticle}
\bauthor{\bsnm{Sun}, \binits{D.}},
\bauthor{\bsnm{Boudouaia}, \binits{A.}},
\bauthor{\bsnm{Zhu}, \binits{C.}},
\bauthor{\bsnm{Li}, \binits{Y.}}:
\batitle{Would chatgpt-facilitated programming mode impact college students’
  programming behaviors, performances, and perceptions? an empirical study}.
\bjtitle{International Journal of Educational Technology in Higher Education}
\bvolume{21}(\bissue{1}),
\bfpage{14}
(\byear{2024})
\end{barticle}
\endbibitem

%%% 9
\bibitem[\protect\citeauthoryear{Denny et~al.}{2023}]{denny2023conversing}
\begin{bchapter}
\bauthor{\bsnm{Denny}, \binits{P.}},
\bauthor{\bsnm{Kumar}, \binits{V.}},
\bauthor{\bsnm{Giacaman}, \binits{N.}}:
\bctitle{Conversing with copilot: Exploring prompt engineering for solving cs1
  problems using natural language}.
In: \bbtitle{Proceedings of the 54th ACM Technical Symposium on Computer
  Science Education V. 1},
pp. \bfpage{1136}--\blpage{1142}
(\byear{2023})
\end{bchapter}
\endbibitem

%%% 10
\bibitem[\protect\citeauthoryear{Finnie-Ansley
  et~al.}{2022}]{finnie-ansley2022robots}
\begin{bchapter}
\bauthor{\bsnm{Finnie-Ansley}, \binits{J.}},
\bauthor{\bsnm{Denny}, \binits{P.}},
\bauthor{\bsnm{Becker}, \binits{B.A.}},
\bauthor{\bsnm{Luxton-Reilly}, \binits{A.}},
\bauthor{\bsnm{Prather}, \binits{J.}}:
\bctitle{The robots are coming: Exploring the implications of openai codex on
  introductory programming}.
In: \bbtitle{Proceedings of the 24th Australasian Computing Education
  Conference}.
\bsertitle{ACE '22},
pp. \bfpage{10}--\blpage{19}.
\bpublisher{Association for Computing Machinery},
\blocation{New York, NY, USA}
(\byear{2022}).
\doiurl{10.1145/3511861.3511863} .
\burl{https://doi.org/10.1145/3511861.3511863}
\end{bchapter}
\endbibitem

%%% 11
\bibitem[\protect\citeauthoryear{Anagnostopoulos}{2023}]{anagnostopoulos2023chatgpt}
\begin{botherref}
\oauthor{\bsnm{Anagnostopoulos}, \binits{C.-N.}}:
Chatgpt impacts in programming education: A recent literature overview that
  debates chatgtp responses.
arXiv preprint arXiv:2309.12348
(2023)
\end{botherref}
\endbibitem

%%% 12
\bibitem[\protect\citeauthoryear{Rajala et~al.}{2023}]{rajala2023call}
\begin{bchapter}
\bauthor{\bsnm{Rajala}, \binits{J.}},
\bauthor{\bsnm{Hukkanen}, \binits{J.}},
\bauthor{\bsnm{Hartikainen}, \binits{M.}},
\bauthor{\bsnm{Niemel{\"a}}, \binits{P.}}:
\bctitle{"call me kiran" -chatgpt as a tutoring chatbot in a computer science
  course}.
In: \bbtitle{Proceedings of the 26th International Academic Mindtrek
  Conference},
pp. \bfpage{83}--\blpage{94}
(\byear{2023})
\end{bchapter}
\endbibitem

%%% 13
\bibitem[\protect\citeauthoryear{Darvishi et~al.}{2024}]{darvishi2024impact}
\begin{barticle}
\bauthor{\bsnm{Darvishi}, \binits{A.}},
\bauthor{\bsnm{Khosravi}, \binits{H.}},
\bauthor{\bsnm{Sadiq}, \binits{S.}},
\bauthor{\bsnm{Ga{\v{s}}evi{\'c}}, \binits{D.}},
\bauthor{\bsnm{Siemens}, \binits{G.}}:
\batitle{Impact of ai assistance on student agency}.
\bjtitle{Computers \& Education}
\bvolume{210},
\bfpage{104967}
(\byear{2024})
\end{barticle}
\endbibitem

%%% 14
\bibitem[\protect\citeauthoryear{Fan et~al.}{2025}]{fan2025beware}
\begin{barticle}
\bauthor{\bsnm{Fan}, \binits{Y.}},
\bauthor{\bsnm{Tang}, \binits{L.}},
\bauthor{\bsnm{Le}, \binits{H.}},
\bauthor{\bsnm{Shen}, \binits{K.}},
\bauthor{\bsnm{Tan}, \binits{S.}},
\bauthor{\bsnm{Zhao}, \binits{Y.}},
\bauthor{\bsnm{Shen}, \binits{Y.}},
\bauthor{\bsnm{Li}, \binits{X.}},
\bauthor{\bsnm{Ga{\v{s}}evi{\'c}}, \binits{D.}}:
\batitle{Beware of metacognitive laziness: Effects of generative artificial
  intelligence on learning motivation, processes, and performance}.
\bjtitle{British Journal of Educational Technology}
\bvolume{56}(\bissue{2}),
\bfpage{489}--\blpage{530}
(\byear{2025})
\end{barticle}
\endbibitem

%%% 15
\bibitem[\protect\citeauthoryear{Skjuve et~al.}{2023}]{skjuve2023user}
\begin{bchapter}
\bauthor{\bsnm{Skjuve}, \binits{M.}},
\bauthor{\bsnm{F{\o}lstad}, \binits{A.}},
\bauthor{\bsnm{Brandtzaeg}, \binits{P.B.}}:
\bctitle{The user experience of chatgpt: Findings from a questionnaire study of
  early users}.
In: \bbtitle{Proceedings of the 5th International Conference on Conversational
  User Interfaces},
pp. \bfpage{1}--\blpage{10}
(\byear{2023})
\end{bchapter}
\endbibitem

%%% 16
\bibitem[\protect\citeauthoryear{Tlili et~al.}{2023}]{tlili2023if}
\begin{barticle}
\bauthor{\bsnm{Tlili}, \binits{A.}},
\bauthor{\bsnm{Shehata}, \binits{B.}},
\bauthor{\bsnm{Adarkwah}, \binits{M.A.}},
\bauthor{\bsnm{Bozkurt}, \binits{A.}},
\bauthor{\bsnm{Hickey}, \binits{D.T.}},
\bauthor{\bsnm{Huang}, \binits{R.}},
\bauthor{\bsnm{Agyemang}, \binits{B.}}:
\batitle{What if the devil is my guardian angel: Chatgpt as a case study of
  using chatbots in education}.
\bjtitle{Smart Learning Environments}
\bvolume{10}(\bissue{1}),
\bfpage{15}
(\byear{2023})
\end{barticle}
\endbibitem

%%% 17
\bibitem[\protect\citeauthoryear{Ma et~al.}{2024}]{ma2024enhancing}
\begin{bchapter}
\bauthor{\bsnm{Ma}, \binits{B.}},
\bauthor{\bsnm{Chen}, \binits{L.}},
\bauthor{\bsnm{Konomi}, \binits{S.}}:
\bctitle{Enhancing programming education with chatgpt: a case study on student
  perceptions and interactions in a python course}.
In: \bbtitle{International Conference on Artificial Intelligence in Education},
pp. \bfpage{113}--\blpage{126}
(\byear{2024}).
\bcomment{Springer}
\end{bchapter}
\endbibitem

%%% 18
\bibitem[\protect\citeauthoryear{Zhang et~al.}{2024}]{zhang2024students}
\begin{bchapter}
\bauthor{\bsnm{Zhang}, \binits{Z.}},
\bauthor{\bsnm{Dong}, \binits{Z.}},
\bauthor{\bsnm{Shi}, \binits{Y.}},
\bauthor{\bsnm{Price}, \binits{T.}},
\bauthor{\bsnm{Matsuda}, \binits{N.}},
\bauthor{\bsnm{Xu}, \binits{D.}}:
\bctitle{Students’ perceptions and preferences of generative artificial
  intelligence feedback for programming}.
In: \bbtitle{Proceedings of the AAAI Conference on Artificial Intelligence},
vol. \bseriesno{38},
pp. \bfpage{23250}--\blpage{23258}
(\byear{2024})
\end{bchapter}
\endbibitem

%%% 19
\bibitem[\protect\citeauthoryear{Loksa et~al.}{2022}]{loksa2022metacognition}
\begin{barticle}
\bauthor{\bsnm{Loksa}, \binits{D.}},
\bauthor{\bsnm{Margulieux}, \binits{L.}},
\bauthor{\bsnm{Becker}, \binits{B.A.}},
\bauthor{\bsnm{Craig}, \binits{M.}},
\bauthor{\bsnm{Denny}, \binits{P.}},
\bauthor{\bsnm{Pettit}, \binits{R.}},
\bauthor{\bsnm{Prather}, \binits{J.}}:
\batitle{Metacognition and self-regulation in programming education: Theories
  and exemplars of use}.
\bjtitle{ACM Transactions on Computing Education (TOCE)}
\bvolume{22}(\bissue{4}),
\bfpage{1}--\blpage{31}
(\byear{2022})
\end{barticle}
\endbibitem

%%% 20
\bibitem[\protect\citeauthoryear{Prather et~al.}{2023}]{prather2023robots}
\begin{bchapter}
\bauthor{\bsnm{Prather}, \binits{J.}},
\bauthor{\bsnm{Denny}, \binits{P.}},
\bauthor{\bsnm{Leinonen}, \binits{J.}},
\bauthor{\bsnm{Becker}, \binits{B.A.}},
\bauthor{\bsnm{Albluwi}, \binits{I.}},
\bauthor{\bsnm{Craig}, \binits{M.}},
\bauthor{\bsnm{Keuning}, \binits{H.}},
\bauthor{\bsnm{Kiesler}, \binits{N.}},
\bauthor{\bsnm{Kohn}, \binits{T.}},
\bauthor{\bsnm{Luxton-Reilly}, \binits{A.}}, \betal:
\bctitle{The robots are here: Navigating the generative ai revolution in
  computing education}.
In: \bbtitle{Proceedings of the 2023 Working Group Reports on Innovation and
  Technology in Computer Science Education},
pp. \bfpage{108}--\blpage{159}
(\byear{2023})
\end{bchapter}
\endbibitem

%%% 21
\bibitem[\protect\citeauthoryear{Zimmerman}{2000}]{zimmerman2000attaining}
\begin{bchapter}
\bauthor{\bsnm{Zimmerman}, \binits{B.J.}}:
\bctitle{Attaining self-regulation: A social cognitive perspective}.
In: \bbtitle{Handbook of Self-regulation},
pp. \bfpage{13}--\blpage{39}.
\bpublisher{Elsevier},
\blocation{San Diego}
(\byear{2000})
\end{bchapter}
\endbibitem

%%% 22
\bibitem[\protect\citeauthoryear{Phung et~al.}{2025}]{phung2025plan}
\begin{bchapter}
\bauthor{\bsnm{Phung}, \binits{T.}},
\bauthor{\bsnm{Choi}, \binits{H.}},
\bauthor{\bsnm{Wu}, \binits{M.}},
\bauthor{\bsnm{Singla}, \binits{A.}},
\bauthor{\bsnm{Brooks}, \binits{C.}}:
\bctitle{Plan more, debug less: Applying metacognitive theory to ai-assisted
  programming education}.
In: \bbtitle{International Conference on Artificial Intelligence in Education},
pp. \bfpage{3}--\blpage{17}
(\byear{2025}).
\bcomment{Springer}
\end{bchapter}
\endbibitem

%%% 23
\bibitem[\protect\citeauthoryear{Finnie-Ansley et~al.}{2023}]{finnie2023my}
\begin{bchapter}
\bauthor{\bsnm{Finnie-Ansley}, \binits{J.}},
\bauthor{\bsnm{Denny}, \binits{P.}},
\bauthor{\bsnm{Luxton-Reilly}, \binits{A.}},
\bauthor{\bsnm{Santos}, \binits{E.A.}},
\bauthor{\bsnm{Prather}, \binits{J.}},
\bauthor{\bsnm{Becker}, \binits{B.A.}}:
\bctitle{My ai wants to know if this will be on the exam: Testing openai’s
  codex on cs2 programming exercises}.
In: \bbtitle{Proceedings of the 25th Australasian Computing Education
  Conference},
pp. \bfpage{97}--\blpage{104}
(\byear{2023})
\end{bchapter}
\endbibitem

%%% 24
\bibitem[\protect\citeauthoryear{Phung et~al.}{2023}]{phung2023generative}
\begin{barticle}
\bauthor{\bsnm{Phung}, \binits{T.}},
\bauthor{\bsnm{P{\u{a}}durean}, \binits{V.-A.}},
\bauthor{\bsnm{Cambronero}, \binits{J.}},
\bauthor{\bsnm{Gulwani}, \binits{S.}},
\bauthor{\bsnm{Kohn}, \binits{T.}},
\bauthor{\bsnm{Majumdar}, \binits{R.}},
\bauthor{\bsnm{Singla}, \binits{A.}},
\bauthor{\bsnm{Soares}, \binits{G.}}:
\batitle{Generative ai for programming education: Benchmarking chatgpt, gpt-4,
  and human tutors}.
\bjtitle{International Journal of Management}
\bvolume{21}(\bissue{2}),
\bfpage{100790}
(\byear{2023})
\end{barticle}
\endbibitem

%%% 25
\bibitem[\protect\citeauthoryear{Sarsa et~al.}{2022}]{sarsa2022automatic}
\begin{bchapter}
\bauthor{\bsnm{Sarsa}, \binits{S.}},
\bauthor{\bsnm{Denny}, \binits{P.}},
\bauthor{\bsnm{Hellas}, \binits{A.}},
\bauthor{\bsnm{Leinonen}, \binits{J.}}:
\bctitle{Automatic generation of programming exercises and code explanations
  using large language models}.
In: \bbtitle{Proceedings of the 2022 ACM Conference on International Computing
  Education Research-Volume 1},
pp. \bfpage{27}--\blpage{43}
(\byear{2022})
\end{bchapter}
\endbibitem

%%% 26
\bibitem[\protect\citeauthoryear{Savelka et~al.}{2023}]{savelka2023thrilled}
\begin{bchapter}
\bauthor{\bsnm{Savelka}, \binits{J.}},
\bauthor{\bsnm{Agarwal}, \binits{A.}},
\bauthor{\bsnm{An}, \binits{M.}},
\bauthor{\bsnm{Bogart}, \binits{C.}},
\bauthor{\bsnm{Sakr}, \binits{M.}}:
\bctitle{Thrilled by your progress! large language models (gpt-4) no longer
  struggle to pass assessments in higher education programming courses}.
In: \bbtitle{Proceedings of the 2023 ACM Conference on International Computing
  Education Research-Volume 1}
(\byear{2023})
\end{bchapter}
\endbibitem

%%% 27
\bibitem[\protect\citeauthoryear{MacNeil et~al.}{2023}]{macneil2023experiences}
\begin{bchapter}
\bauthor{\bsnm{MacNeil}, \binits{S.}},
\bauthor{\bsnm{Tran}, \binits{A.}},
\bauthor{\bsnm{Hellas}, \binits{A.}},
\bauthor{\bsnm{Kim}, \binits{J.}},
\bauthor{\bsnm{Sarsa}, \binits{S.}},
\bauthor{\bsnm{Denny}, \binits{P.}},
\bauthor{\bsnm{Bernstein}, \binits{S.}},
\bauthor{\bsnm{Leinonen}, \binits{J.}}:
\bctitle{Experiences from using code explanations generated by large language
  models in a web software development e-book}.
In: \bbtitle{Proceedings of the 54th ACM Technical Symposium on Computer
  Science Education V. 1},
pp. \bfpage{931}--\blpage{937}
(\byear{2023})
\end{bchapter}
\endbibitem

%%% 28
\bibitem[\protect\citeauthoryear{Leinonen et~al.}{2023}]{leinonen2023comparing}
\begin{bchapter}
\bauthor{\bsnm{Leinonen}, \binits{J.}},
\bauthor{\bsnm{Denny}, \binits{P.}},
\bauthor{\bsnm{MacNeil}, \binits{S.}},
\bauthor{\bsnm{Sarsa}, \binits{S.}},
\bauthor{\bsnm{Bernstein}, \binits{S.}},
\bauthor{\bsnm{Kim}, \binits{J.}},
\bauthor{\bsnm{Tran}, \binits{A.}},
\bauthor{\bsnm{Hellas}, \binits{A.}}:
\bctitle{Comparing code explanations created by students and large language
  models}.
In: \bbtitle{Proceedings of the 2023 Conference on Innovation and Technology in
  Computer Science Education V. 1},
pp. \bfpage{124}--\blpage{130}
(\byear{2023})
\end{bchapter}
\endbibitem

%%% 29
\bibitem[\protect\citeauthoryear{Pankiewicz and
  Baker}{2023}]{pankiewicz2023large}
\begin{botherref}
\oauthor{\bsnm{Pankiewicz}, \binits{M.}},
\oauthor{\bsnm{Baker}, \binits{R.S.}}:
Large language models (gpt) for automating feedback on programming assignments.
arXiv preprint arXiv:2307.00150
(2023)
\end{botherref}
\endbibitem

%%% 30
\bibitem[\protect\citeauthoryear{Biswas}{2023}]{biswas2023role}
\begin{barticle}
\bauthor{\bsnm{Biswas}, \binits{S.}}:
\batitle{Role of chatgpt in computer programming.: Chatgpt in computer
  programming.}
\bjtitle{Mesopotamian Journal of Computer Science}
\bvolume{2023},
\bfpage{8}--\blpage{16}
(\byear{2023})
\end{barticle}
\endbibitem

%%% 31
\bibitem[\protect\citeauthoryear{Humble et~al.}{2023}]{humble2023cheaters}
\begin{botherref}
\oauthor{\bsnm{Humble}, \binits{N.}},
\oauthor{\bsnm{Boustedt}, \binits{J.}},
\oauthor{\bsnm{Holmgren}, \binits{H.}},
\oauthor{\bsnm{Milutinovic}, \binits{G.}},
\oauthor{\bsnm{Seipel}, \binits{S.}},
\oauthor{\bsnm{{\"O}stberg}, \binits{A.-S.}}:
Cheaters or ai-enhanced learners: Consequences of chatgpt for programming
  education.
Electronic Journal of e-Learning,
00--00
(2023)
\end{botherref}
\endbibitem

%%% 32
\bibitem[\protect\citeauthoryear{Shoufan}{2023}]{shoufan2023exploring}
\begin{botherref}
\oauthor{\bsnm{Shoufan}, \binits{A.}}:
Exploring students’ perceptions of chatgpt: Thematic analysis and follow-up
  survey.
IEEE Access
(2023)
\end{botherref}
\endbibitem

%%% 33
\bibitem[\protect\citeauthoryear{Ma et~al.}{2024}]{ma2024exploring}
\begin{bchapter}
\bauthor{\bsnm{Ma}, \binits{B.}},
\bauthor{\bsnm{Chen}, \binits{L.}},
\bauthor{\bsnm{Konomi}, \binits{S.}}:
\bctitle{Exploring student perception and interaction using chatgpt in
  programming education}.
In: \bbtitle{21st International Conference on Cognition and Exploratory
  Learning in the Digital Age, CELDA 2024},
pp. \bfpage{35}--\blpage{42}
(\byear{2024}).
\bcomment{IADIS Press}
\end{bchapter}
\endbibitem

%%% 34
\bibitem[\protect\citeauthoryear{Kasneci et~al.}{2023}]{kasneci2023chatgpt}
\begin{barticle}
\bauthor{\bsnm{Kasneci}, \binits{E.}},
\bauthor{\bsnm{Se{\ss}ler}, \binits{K.}},
\bauthor{\bsnm{K{\"u}chemann}, \binits{S.}},
\bauthor{\bsnm{Bannert}, \binits{M.}},
\bauthor{\bsnm{Dementieva}, \binits{D.}},
\bauthor{\bsnm{Fischer}, \binits{F.}},
\bauthor{\bsnm{Gasser}, \binits{U.}},
\bauthor{\bsnm{Groh}, \binits{G.}},
\bauthor{\bsnm{G{\"u}nnemann}, \binits{S.}},
\bauthor{\bsnm{H{\"u}llermeier}, \binits{E.}}, \betal:
\batitle{Chatgpt for good? on opportunities and challenges of large language
  models for education}.
\bjtitle{Learning and Individual Differences}
\bvolume{103},
\bfpage{102274}
(\byear{2023})
\end{barticle}
\endbibitem

%%% 35
\bibitem[\protect\citeauthoryear{Gao et~al.}{2022}]{gao2022who}
\begin{bchapter}
\bauthor{\bsnm{Gao}, \binits{Z.}},
\bauthor{\bsnm{Heckman}, \binits{S.}},
\bauthor{\bsnm{Lynch}, \binits{C.}}:
\bctitle{Who uses office hours? a comparison of in-person and virtual office
  hours utilization}.
In: \bbtitle{Proceedings of the 53rd ACM Technical Symposium on Computer
  Science Education - Volume 1}.
\bsertitle{SIGCSE 2022},
pp. \bfpage{300}--\blpage{306}.
\bpublisher{Association for Computing Machinery},
\blocation{New York, NY, USA}
(\byear{2022}).
\doiurl{10.1145/3478431.3499334} .
\burl{https://doi.org/10.1145/3478431.3499334}
\end{bchapter}
\endbibitem

%%% 36
\bibitem[\protect\citeauthoryear{Denny et~al.}{2023}]{denny2023computing}
\begin{botherref}
\oauthor{\bsnm{Denny}, \binits{P.}},
\oauthor{\bsnm{Prather}, \binits{J.}},
\oauthor{\bsnm{Becker}, \binits{B.A.}},
\oauthor{\bsnm{Finnie-Ansley}, \binits{J.}},
\oauthor{\bsnm{Hellas}, \binits{A.}},
\oauthor{\bsnm{Leinonen}, \binits{J.}},
\oauthor{\bsnm{Luxton-Reilly}, \binits{A.}},
\oauthor{\bsnm{Reeves}, \binits{B.N.}},
\oauthor{\bsnm{Santos}, \binits{E.A.}},
\oauthor{\bsnm{Sarsa}, \binits{S.}}:
Computing education in the era of generative ai.
arXiv preprint arXiv:2306.02608
(2023)
\end{botherref}
\endbibitem

%%% 37
\bibitem[\protect\citeauthoryear{Chan and Lee}{2023}]{chan2023ai}
\begin{barticle}
\bauthor{\bsnm{Chan}, \binits{C.K.Y.}},
\bauthor{\bsnm{Lee}, \binits{K.K.}}:
\batitle{The ai generation gap: Are gen z students more interested in adopting
  generative ai such as chatgpt in teaching and learning than their gen x and
  millennial generation teachers?}
\bjtitle{Smart learning environments}
\bvolume{10}(\bissue{1}),
\bfpage{60}
(\byear{2023})
\end{barticle}
\endbibitem

%%% 38
\bibitem[\protect\citeauthoryear{Amani et~al.}{2023}]{amani2023generative}
\begin{botherref}
\oauthor{\bsnm{Amani}, \binits{S.}},
\oauthor{\bsnm{White}, \binits{L.}},
\oauthor{\bsnm{Balart}, \binits{T.}},
\oauthor{\bsnm{Arora}, \binits{L.}},
\oauthor{\bsnm{Shryock}, \binits{K.J.}},
\oauthor{\bsnm{Brumbelow}, \binits{K.}},
\oauthor{\bsnm{Watson}, \binits{K.L.}}:
Generative ai perceptions: A survey to measure the perceptions of faculty,
  staff, and students on generative ai tools in academia.
arXiv preprint arXiv:2304.14415
(2023)
\end{botherref}
\endbibitem

%%% 39
\bibitem[\protect\citeauthoryear{Lau and Guo}{2023}]{lau2023ban}
\begin{bchapter}
\bauthor{\bsnm{Lau}, \binits{S.}},
\bauthor{\bsnm{Guo}, \binits{P.J.}}:
\bctitle{From" ban it till we understand it" to" resistance is futile": How
  university programming instructors plan to adapt as more students use ai code
  generation and explanation tools such as chatgpt and github copilot}.
In: \bbtitle{Proceedings of the 2023 ACM Conference on International Computing
  Education Research-Volume 1}
(\byear{2023})
\end{bchapter}
\endbibitem

%%% 40
\bibitem[\protect\citeauthoryear{Becker et~al.}{2023}]{becker2023programming}
\begin{bchapter}
\bauthor{\bsnm{Becker}, \binits{B.A.}},
\bauthor{\bsnm{Denny}, \binits{P.}},
\bauthor{\bsnm{Finnie-Ansley}, \binits{J.}},
\bauthor{\bsnm{Luxton-Reilly}, \binits{A.}},
\bauthor{\bsnm{Prather}, \binits{J.}},
\bauthor{\bsnm{Santos}, \binits{E.A.}}:
\bctitle{Programming is hard-or at least it used to be: Educational
  opportunities and challenges of ai code generation}.
In: \bbtitle{Proceedings of the 54th ACM Technical Symposium on Computer
  Science Education V. 1},
pp. \bfpage{500}--\blpage{506}
(\byear{2023})
\end{bchapter}
\endbibitem

%%% 41
\bibitem[\protect\citeauthoryear{Amoozadeh et~al.}{2024}]{amoozadeh2024student}
\begin{botherref}
\oauthor{\bsnm{Amoozadeh}, \binits{M.}},
\oauthor{\bsnm{Nam}, \binits{D.}},
\oauthor{\bsnm{Prol}, \binits{D.}},
\oauthor{\bsnm{Alfageeh}, \binits{A.}},
\oauthor{\bsnm{Prather}, \binits{J.}},
\oauthor{\bsnm{Hilton}, \binits{M.}},
\oauthor{\bsnm{Ragavan}, \binits{S.S.}},
\oauthor{\bsnm{Alipour}, \binits{M.A.}}:
Student-AI Interaction: A Case Study of CS1 Students
(2024).
\doiurl{10.48550/arXiv.2407.00305} .
\url{https://arxiv.org/abs/2407.00305}
\end{botherref}
\endbibitem

%%% 42
\bibitem[\protect\citeauthoryear{Scholl et~al.}{2024}]{scholl2024chatprotocols}
\begin{bchapter}
\bauthor{\bsnm{Scholl}, \binits{A.}},
\bauthor{\bsnm{Schiffner}, \binits{D.}},
\bauthor{\bsnm{Kiesler}, \binits{N.}}:
\bctitle{Analyzing chat protocols of novice programmers solving introductory
  programming tasks with chatgpt}.
In: \bbtitle{DELFI 2024}
(\byear{2024}).
\doiurl{10.18420/delfi2024_05}
\end{bchapter}
\endbibitem

%%% 43
\bibitem[\protect\citeauthoryear{Scholl and Kiesler}{2024}]{scholl2024novice}
\begin{bchapter}
\bauthor{\bsnm{Scholl}, \binits{A.}},
\bauthor{\bsnm{Kiesler}, \binits{N.}}:
\bctitle{How novice programmers use and experience chatgpt when solving
  programming exercises in an introductory course}.
In: \bbtitle{2024 IEEE Frontiers in Education Conference (FIE)}.
\bpublisher{IEEE},
\blocation{Piscataway, NJ, USA}
(\byear{2024}).
\doiurl{10.1109/FIE61694.2024.10893442}
\end{bchapter}
\endbibitem

%%% 44
\bibitem[\protect\citeauthoryear{Sheese et~al.}{2024}]{sheese2024patterns}
\begin{bchapter}
\bauthor{\bsnm{Sheese}, \binits{B.}},
\bauthor{\bsnm{Liffiton}, \binits{M.H.}},
\bauthor{\bsnm{Savelka}, \binits{J.}},
\bauthor{\bsnm{Denny}, \binits{P.}}:
\bctitle{Patterns of student help-seeking when using a large language
  model-powered programming assistant}.
In: \bbtitle{Proceedings of the 26th Australasian Computing Education
  Conference},
pp. \bfpage{49}--\blpage{57}.
\bpublisher{Association for Computing Machinery},
\blocation{New York, NY, USA}
(\byear{2024}).
\doiurl{10.1145/3636243.3636249}
\end{bchapter}
\endbibitem

%%% 45
\bibitem[\protect\citeauthoryear{Viberg et~al.}{2025}]{viberg2025chatting}
\begin{bchapter}
\bauthor{\bsnm{Viberg}, \binits{O.}},
\bauthor{\bsnm{Wong}, \binits{J.}},
\bauthor{\bsnm{Feldman-Maggor}, \binits{Y.}},
\bauthor{\bsnm{Dunder}, \binits{N.}},
\bauthor{\bsnm{Epp}, \binits{C.D.}}:
\bctitle{Chatting with code: Exploring llms as learning partners in programming
  education}.
In: \bbtitle{International Conference on Artificial Intelligence in Education},
pp. \bfpage{453}--\blpage{461}
(\byear{2025}).
\bcomment{Springer}
\end{bchapter}
\endbibitem

%%% 46
\bibitem[\protect\citeauthoryear{Yang et~al.}{2024}]{yang2024debugging}
\begin{bchapter}
\bauthor{\bsnm{Yang}, \binits{S.}},
\bauthor{\bsnm{Zhao}, \binits{H.}},
\bauthor{\bsnm{Xu}, \binits{Y.}},
\bauthor{\bsnm{Brennan}, \binits{K.}},
\bauthor{\bsnm{Schneider}, \binits{B.}}:
\bctitle{Debugging with an ai tutor: Investigating novice help-seeking
  behaviors and perceived learning}.
In: \bbtitle{Proceedings of the 2024 ACM Conference on International Computing
  Education Research}.
\bpublisher{Association for Computing Machinery},
\blocation{New York, NY, USA}
(\byear{2024}).
\doiurl{10.1145/3632620.3671092}
\end{bchapter}
\endbibitem

%%% 47
\bibitem[\protect\citeauthoryear{Liffiton et~al.}{2023}]{liffiton2023codehelp}
\begin{botherref}
\oauthor{\bsnm{Liffiton}, \binits{M.}},
\oauthor{\bsnm{Sheese}, \binits{B.}},
\oauthor{\bsnm{Savelka}, \binits{J.}},
\oauthor{\bsnm{Denny}, \binits{P.}}:
CodeHelp: Using Large Language Models with Guardrails for Scalable Support in
  Programming Classes
(2023)
\end{botherref}
\endbibitem

%%% 48
\bibitem[\protect\citeauthoryear{Prather et~al.}{2024}]{prather2024widening}
\begin{bchapter}
\bauthor{\bsnm{Prather}, \binits{J.}},
\bauthor{\bsnm{Reeves}, \binits{B.}},
\bauthor{\bsnm{Leinonen}, \binits{J.}},
\bauthor{\bsnm{MacNeil}, \binits{S.}},
\bauthor{\bsnm{Randrianasolo}, \binits{A.}},
\bauthor{\bsnm{Becker}, \binits{B.A.}},
\bauthor{\bsnm{Kimmel}, \binits{B.}},
\bauthor{\bsnm{Wright}, \binits{J.}},
\bauthor{\bsnm{Briggs}, \binits{B.}}:
\bctitle{The widening gap: The benefits and harms of generative ai for novice
  programmers}.
In: \bbtitle{Proceedings of the 2024 ACM Conference on International Computing
  Education Research}.
\bpublisher{Association for Computing Machinery},
\blocation{New York, NY, USA}
(\byear{2024})
\end{bchapter}
\endbibitem

%%% 49
\bibitem[\protect\citeauthoryear{Ebrahimi et~al.}{2006}]{ebrahimi2006taxonomy}
\begin{barticle}
\bauthor{\bsnm{Ebrahimi}, \binits{A.}},
\bauthor{\bsnm{Kopec}, \binits{D.}},
\bauthor{\bsnm{Schweikert}, \binits{C.}}:
\batitle{Taxonomy of novice programming error patterns with plan, web, and
  object solutions}.
\bjtitle{ACM Computing Surveys}
\bvolume{38}(\bissue{2}),
\bfpage{1}--\blpage{24}
(\byear{2006})
\end{barticle}
\endbibitem

%%% 50
\bibitem[\protect\citeauthoryear{Hao and Liu}{2025}]{hao2025towards}
\begin{bchapter}
\bauthor{\bsnm{Hao}, \binits{Q.}},
\bauthor{\bsnm{Liu}, \binits{R.}}:
\bctitle{Towards integrating behavior-driven development in mobile development:
  An experience report}.
In: \bbtitle{Proceedings of the 56th ACM Technical Symposium on Computer
  Science Education V. 1},
pp. \bfpage{450}--\blpage{456}
(\byear{2025})
\end{bchapter}
\endbibitem

%%% 51
\bibitem[\protect\citeauthoryear{Park and Cheon}{2025}]{park2025exploring}
\begin{barticle}
\bauthor{\bsnm{Park}, \binits{E.}},
\bauthor{\bsnm{Cheon}, \binits{J.}}:
\batitle{Exploring debugging challenges and strategies using structural topic
  model: A comparative analysis of high and low-performing students}.
\bjtitle{Journal of Educational Computing Research}
\bvolume{62}(\bissue{8}),
\bfpage{1884}--\blpage{1906}
(\byear{2025})
\end{barticle}
\endbibitem

%%% 52
\bibitem[\protect\citeauthoryear{Saliba et~al.}{2024}]{saliba2024learning}
\begin{bchapter}
\bauthor{\bsnm{Saliba}, \binits{L.}},
\bauthor{\bsnm{Shioji}, \binits{E.}},
\bauthor{\bsnm{Oliveira}, \binits{E.}},
\bauthor{\bsnm{Cohney}, \binits{S.}},
\bauthor{\bsnm{Qi}, \binits{J.}}:
\bctitle{Learning with style: Improving student code-style through better
  automated feedback}.
In: \bbtitle{Proceedings of the 55th ACM Technical Symposium on Computer
  Science Education V. 1},
pp. \bfpage{1175}--\blpage{1181}
(\byear{2024})
\end{bchapter}
\endbibitem

%%% 53
\bibitem[\protect\citeauthoryear{Choi et~al.}{2023}]{choi2023benefit}
\begin{barticle}
\bauthor{\bsnm{Choi}, \binits{H.}},
\bauthor{\bsnm{Jovanovic}, \binits{J.}},
\bauthor{\bsnm{Poquet}, \binits{O.}},
\bauthor{\bsnm{Brooks}, \binits{C.}},
\bauthor{\bsnm{Joksimovi{\'c}}, \binits{S.}},
\bauthor{\bsnm{Williams}, \binits{J.J.}}:
\batitle{The benefit of reflection prompts for encouraging learning with hints
  in an online programming course}.
\bjtitle{The Internet and Higher Education}
\bvolume{58},
\bfpage{100903}
(\byear{2023})
\end{barticle}
\endbibitem

%%% 54
\bibitem[\protect\citeauthoryear{Karaoglan~Yilmaz and
  Yilmaz}{2022}]{karaoglan2022learning}
\begin{barticle}
\bauthor{\bsnm{Karaoglan~Yilmaz}, \binits{F.G.}},
\bauthor{\bsnm{Yilmaz}, \binits{R.}}:
\batitle{Learning analytics intervention improves students’ engagement in
  online learning}.
\bjtitle{Technology, Knowledge and Learning}
\bvolume{27}(\bissue{2}),
\bfpage{449}--\blpage{460}
(\byear{2022})
\end{barticle}
\endbibitem

%%% 55
\bibitem[\protect\citeauthoryear{Cheng et~al.}{2024}]{cheng2024exploring}
\begin{barticle}
\bauthor{\bsnm{Cheng}, \binits{G.}},
\bauthor{\bsnm{Zou}, \binits{D.}},
\bauthor{\bsnm{Xie}, \binits{H.}},
\bauthor{\bsnm{Wang}, \binits{F.L.}}:
\batitle{Exploring differences in self-regulated learning strategy use between
  high-and low-performing students in introductory programming: An analysis of
  eye-tracking and retrospective think-aloud data from program comprehension}.
\bjtitle{Computers \& Education}
\bvolume{208},
\bfpage{104948}
(\byear{2024})
\end{barticle}
\endbibitem

%%% 56
\bibitem[\protect\citeauthoryear{Ma et~al.}{2025}]{ma2025scaffolding}
\begin{botherref}
\oauthor{\bsnm{Ma}, \binits{B.}},
\oauthor{\bsnm{Li}, \binits{H.}},
\oauthor{\bsnm{Li}, \binits{G.}},
\oauthor{\bsnm{Chen}, \binits{L.}},
\oauthor{\bsnm{Tang}, \binits{C.}},
\oauthor{\bsnm{Xie}, \binits{Y.}},
\oauthor{\bsnm{Gu}, \binits{C.}},
\oauthor{\bsnm{Shimada}, \binits{A.}},
\oauthor{\bsnm{Konomi}, \binits{S.}}:
Scaffolding metacognition in programming education: Understanding student-ai
  interactions and design implications.
arXiv preprint arXiv:2511.04144
(2025)
\end{botherref}
\endbibitem

%%% 57
\bibitem[\protect\citeauthoryear{Chen et~al.}{2025}]{chen2025unpacking}
\begin{barticle}
\bauthor{\bsnm{Chen}, \binits{A.}},
\bauthor{\bsnm{Xiang}, \binits{M.}},
\bauthor{\bsnm{Zhou}, \binits{J.}},
\bauthor{\bsnm{Jia}, \binits{J.}},
\bauthor{\bsnm{Shang}, \binits{J.}},
\bauthor{\bsnm{Li}, \binits{X.}},
\bauthor{\bsnm{Ga{\v{s}}evi{\'c}}, \binits{D.}},
\bauthor{\bsnm{Fan}, \binits{Y.}}:
\batitle{Unpacking help-seeking process through multimodal learning analytics:
  A comparative study of chatgpt vs human expert}.
\bjtitle{Computers \& Education}
\bvolume{226},
\bfpage{105198}
(\byear{2025})
\end{barticle}
\endbibitem

%%% 58
\bibitem[\protect\citeauthoryear{Kazemitabaar
  et~al.}{2023}]{kazemitabaar2023studying}
\begin{bchapter}
\bauthor{\bsnm{Kazemitabaar}, \binits{M.}},
\bauthor{\bsnm{Chow}, \binits{J.}},
\bauthor{\bsnm{Ma}, \binits{C.K.T.}},
\bauthor{\bsnm{Ericson}, \binits{B.J.}},
\bauthor{\bsnm{Weintrop}, \binits{D.}},
\bauthor{\bsnm{Grossman}, \binits{T.}}:
\bctitle{Studying the effect of ai code generators on supporting novice
  learners in introductory programming}.
In: \bbtitle{Proceedings of the 2023 CHI Conference on Human Factors in
  Computing Systems},
pp. \bfpage{1}--\blpage{23}
(\byear{2023})
\end{bchapter}
\endbibitem

%%% 59
\bibitem[\protect\citeauthoryear{Penney et~al.}{2025}]{penney2025preferences}
\begin{bchapter}
\bauthor{\bsnm{Penney}, \binits{J.}},
\bauthor{\bsnm{Acharya}, \binits{P.}},
\bauthor{\bsnm{Hilbert}, \binits{P.}},
\bauthor{\bsnm{Parekh}, \binits{P.}},
\bauthor{\bsnm{Sarma}, \binits{A.}},
\bauthor{\bsnm{Steinmacher}, \binits{I.}},
\bauthor{\bsnm{Gerosa}, \binits{M.A.}}:
\bctitle{Understanding programming students' help-seeking preferences in the
  era of generative ai}.
In: \bbtitle{CompEd 2025 -- Proceedings of the ACM Global Computing Education
  Conference 2025},
pp. \bfpage{15}--\blpage{21}.
\bpublisher{Association for Computing Machinery},
\blocation{New York, NY, USA}
(\byear{2025}).
\doiurl{10.1145/3736181.3747165}
\end{bchapter}
\endbibitem

%%% 60
\bibitem[\protect\citeauthoryear{Li and Ma}{2025}]{li2025coderunner}
\begin{bchapter}
\bauthor{\bsnm{Li}, \binits{H.}},
\bauthor{\bsnm{Ma}, \binits{B.}}:
\bctitle{Coderunner agent: Integrating ai feedback and self-regulated learning
  to support programming education}.
In: \bbtitle{International Conference on Computers in Education}
(\byear{2025})
\end{bchapter}
\endbibitem

%%% 61
\bibitem[\protect\citeauthoryear{Kazemitabaar
  et~al.}{2023}]{kazemitabaar2023how_novices_use_llm_code}
\begin{bchapter}
\bauthor{\bsnm{Kazemitabaar}, \binits{M.}},
\bauthor{\bsnm{Hou}, \binits{X.}},
\bauthor{\bsnm{Henley}, \binits{A.}},
\bauthor{\bsnm{Ericson}, \binits{B.}},
\bauthor{\bsnm{Weintrop}, \binits{D.}},
\bauthor{\bsnm{Grossman}, \binits{T.}}:
\bctitle{How novices use llm-based code generators to solve cs1 coding tasks in
  a self-paced learning environment}.
In: \bbtitle{Proceedings of the 23rd Koli Calling International Conference on
  Computing Education Research}
(\byear{2023})
\end{bchapter}
\endbibitem

%%% 62
\bibitem[\protect\citeauthoryear{Brandt et~al.}{2009}]{brandt2009two}
\begin{bchapter}
\bauthor{\bsnm{Brandt}, \binits{J.}},
\bauthor{\bsnm{Guo}, \binits{P.J.}},
\bauthor{\bsnm{Lewenstein}, \binits{J.}},
\bauthor{\bsnm{Dontcheva}, \binits{M.}},
\bauthor{\bsnm{Klemmer}, \binits{S.R.}}:
\bctitle{Two studies of opportunistic programming: interleaving web foraging,
  learning, and writing code}.
In: \bbtitle{Proceedings of the SIGCHI Conference on Human Factors in Computing
  Systems},
pp. \bfpage{1589}--\blpage{1598}
(\byear{2009})
\end{bchapter}
\endbibitem

%%% 63
\bibitem[\protect\citeauthoryear{Karabenick and
  Berger}{2013}]{karabenick2013help}
\begin{bchapter}
\bauthor{\bsnm{Karabenick}, \binits{S.A.}},
\bauthor{\bsnm{Berger}, \binits{J.-L.}}:
\bctitle{Help seeking as a self-regulated learning strategy}.
In: \beditor{\bsnm{Bembenutty}, \binits{H.}},
\beditor{\bsnm{Cleary}, \binits{T.J.}},
\beditor{\bsnm{Kitsantas}, \binits{A.}} (eds.)
\bbtitle{Applications of Self-Regulated Learning Across Diverse Disciplines: A
  Tribute to Barry J. Zimmerman},
pp. \bfpage{237}--\blpage{261}.
\bpublisher{Information Age Publishing},
\blocation{Charlotte, NC, USA}
(\byear{2013}).
\doiurl{10.1108/978-1-62396-134-320251009}
\end{bchapter}
\endbibitem

%%% 64
\bibitem[\protect\citeauthoryear{Prasad and Sane}{2024}]{prasad2024self}
\begin{bchapter}
\bauthor{\bsnm{Prasad}, \binits{P.}},
\bauthor{\bsnm{Sane}, \binits{A.}}:
\bctitle{A self-regulated learning framework using generative ai and its
  application in cs educational intervention design}.
In: \bbtitle{Proceedings of the 55th ACM Technical Symposium on Computer
  Science Education V. 1},
pp. \bfpage{1070}--\blpage{1076}
(\byear{2024})
\end{bchapter}
\endbibitem

%%% 65
\bibitem[\protect\citeauthoryear{Silva et~al.}{2024}]{silva2024learning}
\begin{barticle}
\bauthor{\bsnm{Silva}, \binits{L.}},
\bauthor{\bsnm{Mendes}, \binits{A.}},
\bauthor{\bsnm{Gomes}, \binits{A.}},
\bauthor{\bsnm{Fortes}, \binits{G.}}:
\batitle{What learning strategies are used by programming students? a
  qualitative study grounded on the self-regulation of learning theory}.
\bjtitle{ACM Transactions on Computing Education}
\bvolume{24}(\bissue{1}),
\bfpage{1}--\blpage{26}
(\byear{2024})
\end{barticle}
\endbibitem

%%% 66
\bibitem[\protect\citeauthoryear{Bingham and
  Witkowsky}{2021}]{bingham2021deductive}
\begin{botherref}
\oauthor{\bsnm{Bingham}, \binits{A.J.}},
\oauthor{\bsnm{Witkowsky}, \binits{P.}}:
Deductive and inductive approaches to qualitative data analysis.
Analyzing and interpreting qualitative data: After the interview,
133--146
(2021)
\end{botherref}
\endbibitem

%%% 67
\bibitem[\protect\citeauthoryear{Miles and
  Huberman}{1994}]{miles1994qualitative}
\begin{bbook}
\bauthor{\bsnm{Miles}, \binits{M.B.}},
\bauthor{\bsnm{Huberman}, \binits{A.M.}}:
\bbtitle{Qualitative Data Analysis: An Expanded Sourcebook}.
\bpublisher{sage},
\blocation{Thousand Oaks, CA, USA}
(\byear{1994})
\end{bbook}
\endbibitem

%%% 68
\bibitem[\protect\citeauthoryear{Neuendorf}{2017}]{neuendorf2017content}
\begin{bbook}
\bauthor{\bsnm{Neuendorf}, \binits{K.A.}}:
\bbtitle{The Content Analysis Guidebook}.
\bpublisher{sage},
\blocation{Thousand Oaks, CA, USA}
(\byear{2017})
\end{bbook}
\endbibitem

%%% 69
\bibitem[\protect\citeauthoryear{Fan et~al.}{2022}]{fan2022improving}
\begin{barticle}
\bauthor{\bsnm{Fan}, \binits{Y.}},
\bauthor{\bsnm{Lim}, \binits{L.}},
\bauthor{\bsnm{Graaf}, \binits{J.}},
\bauthor{\bsnm{Kilgour}, \binits{J.}},
\bauthor{\bsnm{Rakovi{\'c}}, \binits{M.}},
\bauthor{\bsnm{Moore}, \binits{J.}},
\bauthor{\bsnm{Molenaar}, \binits{I.}},
\bauthor{\bsnm{Bannert}, \binits{M.}},
\bauthor{\bsnm{Ga{\v{s}}evi{\'c}}, \binits{D.}}:
\batitle{Improving the measurement of self-regulated learning using
  multi-channel data}.
\bjtitle{Metacognition and Learning}
\bvolume{17}(\bissue{3}),
\bfpage{1025}--\blpage{1055}
(\byear{2022})
\end{barticle}
\endbibitem

%%% 70
\bibitem[\protect\citeauthoryear{Aleven et~al.}{2006}]{aleven2006toward}
\begin{barticle}
\bauthor{\bsnm{Aleven}, \binits{V.}},
\bauthor{\bsnm{McLaren}, \binits{B.}},
\bauthor{\bsnm{Roll}, \binits{I.}},
\bauthor{\bsnm{Koedinger}, \binits{K.}}:
\batitle{Toward meta-cognitive tutoring: A model of help seeking with a
  cognitive tutor}.
\bjtitle{International Journal of Artificial Intelligence in Education}
\bvolume{16}(\bissue{2}),
\bfpage{101}--\blpage{128}
(\byear{2006})
\end{barticle}
\endbibitem

%%% 71
\bibitem[\protect\citeauthoryear{Aleven et~al.}{2016}]{aleven2016help}
\begin{barticle}
\bauthor{\bsnm{Aleven}, \binits{V.}},
\bauthor{\bsnm{Roll}, \binits{I.}},
\bauthor{\bsnm{McLaren}, \binits{B.M.}},
\bauthor{\bsnm{Koedinger}, \binits{K.R.}}:
\batitle{Help helps, but only so much: Research on help seeking with
  intelligent tutoring systems}.
\bjtitle{International Journal of Artificial Intelligence in Education}
\bvolume{26}(\bissue{1}),
\bfpage{205}--\blpage{223}
(\byear{2016})
\end{barticle}
\endbibitem

%%% 72
\bibitem[\protect\citeauthoryear{Ma et~al.}{2026}]{ma2026design}
\begin{botherref}
\oauthor{\bsnm{Ma}, \binits{B.}},
\oauthor{\bsnm{Xie}, \binits{Y.}},
\oauthor{\bsnm{Li}, \binits{H.}},
\oauthor{\bsnm{Li}, \binits{G.}},
\oauthor{\bsnm{Chen}, \binits{L.}},
\oauthor{\bsnm{Shimada}, \binits{A.}},
\oauthor{\bsnm{Konomi}, \binits{S.}}:
Design implications for student and educator needs in ai-supported programming
  learning tools.
arXiv preprint arXiv:2603.22673
(2026)
\end{botherref}
\endbibitem

%%% 73
\bibitem[\protect\citeauthoryear{Chen et~al.}{2024}]{chen2024stugptviz}
\begin{barticle}
\bauthor{\bsnm{Chen}, \binits{Z.}},
\bauthor{\bsnm{Wang}, \binits{J.}},
\bauthor{\bsnm{Xia}, \binits{M.}},
\bauthor{\bsnm{Shigyo}, \binits{K.}},
\bauthor{\bsnm{Liu}, \binits{D.}},
\bauthor{\bsnm{Zhang}, \binits{R.}},
\bauthor{\bsnm{Qu}, \binits{H.}}:
\batitle{Stugptviz: A visual analytics approach to understand student-chatgpt
  interactions}.
\bjtitle{IEEE Transactions on Visualization and Computer Graphics}
\bvolume{31}(\bissue{1}),
\bfpage{908}--\blpage{918}
(\byear{2024})
\end{barticle}
\endbibitem

\end{thebibliography}

\end{document}